\documentclass{article}

\usepackage{microtype}
\usepackage{graphicx}
\usepackage[raggedright]{subfigure}
\usepackage{booktabs} 

\usepackage{float,amsmath,hyperref}

\usepackage[accepted]{icml2020_style/icml2020}

\usepackage{amsfonts,amsthm,amssymb}
\usepackage{bm}
\usepackage{sty/notation}
\usepackage{enumitem,array}
\usepackage{dsfont}
\usepackage{multicol}
\usepackage{subfiles}

\DeclareMathOperator*{\argmax}{argmax}
\DeclareMathOperator*{\argmin}{argmin}

\DeclareMathOperator{\ex}{\mathbb E}

\DeclareMathOperator{\R}{\mathbb R}
\DeclareMathOperator{\N}{\mathbb N}

\DeclareMathOperator{\UCB}{UCB}

\DeclareMathOperator{\indicator}{\mathds{1}}
\DeclareMathOperator{\bigO}{\mathcal{O}}

\let\inf\undefined
\let\min\undefined
\let\max\undefined
\DeclareMathOperator*{\inf}{\vphantom{sup}inf}
\DeclareMathOperator*{\min}{\vphantom{sup}min}
\DeclareMathOperator*{\max}{\vphantom{sup}max}

\DeclareBoldMathCommand{\vmu}{\mu}
\DeclareBoldMathCommand{\vlambda}{\lambda}
\DeclareBoldMathCommand{\vtheta}{\theta}
\DeclareBoldMathCommand{\vxi}{\xi}
\DeclareBoldMathCommand{\veta}{\eta}
\DeclareBoldMathCommand{\vzeta}{\zeta}
\DeclareBoldMathCommand{\ones}{1}
\DeclareBoldMathCommand{\zeros}{0}
\DeclareBoldMathCommand{\A}{A}
\DeclareBoldMathCommand{\I}{I}
\DeclareBoldMathCommand{\v}{v}
\DeclareBoldMathCommand{\w}{w}
\DeclareBoldMathCommand{\q}{q}
\DeclareBoldMathCommand{\p}{p}
\DeclareBoldMathCommand{\e}{e}
\DeclareBoldMathCommand{\a}{a}
\DeclareBoldMathCommand{\u}{u}
\DeclareBoldMathCommand{\vN}{N}
\DeclareBoldMathCommand{\Z}{Z}
\DeclareBoldMathCommand{\z}{z}
\DeclareBoldMathCommand{\c}{c}
\DeclareBoldMathCommand{\x}{x}
\DeclareBoldMathCommand{\vDelta}{\Delta}

\let\top\intercal

\newtheorem{lemma}{Lemma}
\newtheorem{theorem}{Theorem}

\theoremstyle{definition}

\newtheorem{definition}{Definition}
\newtheorem{assumption}{Assumption}

\icmltitlerunning{Structure Adaptive Algorithms for Stochastic Bandits}

\begin{document}

\twocolumn[
\icmltitle{Structure Adaptive Algorithms for Stochastic Bandits}




\begin{icmlauthorlist}
\icmlauthor{R\'emy Degenne}{inria}
\icmlauthor{Han Shao}{a}
\icmlauthor{Wouter M. Koolen}{b}
\end{icmlauthorlist}

\icmlaffiliation{inria}{INRIA - DIENS - PSL Research University, Paris, France}
\icmlaffiliation{a}{Toyota Technological Institute at Chicago}
\icmlaffiliation{b}{Centrum Wiskunde \& Informatica}

\icmlcorrespondingauthor{R\'emy Degenne}{\texttt{remydegenne@gmail.com}}

\icmlkeywords{Machine Learning, ICML}

\vskip 0.3in
]



\printAffiliationsAndNotice{}  

\begin{abstract}
We study reward maximisation in a wide class of structured stochastic multi-armed bandit problems, where the mean rewards of arms satisfy some given structural constraints, e.g. linear, unimodal, sparse, etc. Our aim is to develop methods that are \emph{flexible} (in that they easily adapt to different structures), \emph{powerful} (in that they perform well empirically and/or provably match instance-dependent lower bounds) and \emph{efficient} in that the per-round computational burden is small.

We develop asymptotically optimal algorithms from instance-dependent lower-bounds using iterative saddle-point solvers. Our approach generalises recent iterative methods for pure exploration to reward maximisation, where a major challenge arises from the estimation of the sub-optimality gaps and their reciprocals. Still we manage to achieve all the above desiderata. Notably, our technique avoids the computational cost of the full-blown saddle point oracle employed by previous work, while at the same time enabling finite-time regret bounds. 

Our experiments reveal that our method successfully leverages the structural assumptions, while its regret is at worst comparable to that of vanilla UCB.
\end{abstract}

\section{Introduction}

Stochastic multi-armed bandits are online learning problems in which an algorithm sequentially chooses among a finite set of actions (it ``pulls an arm'') and receives a stochastic reward in return. The goal is to obtain the maximal cumulative reward over time. Introduced by \citet{thompson1933likelihood}, bandits have been the subject of intense study for many years now, starting with asymptotic results in the 80s and 90s \cite{lai1985asymptotically,GravesLai97} and moving to the finite time analysis of algorithms since the beginning of the new millennium, notably with the introduction of strategies based on the Optimism in Face of Uncertainty principle \cite{agrawal1995sample,auer2002finite}. For an overview of the field, we point the reader to \citep{bubeck2012regret, lattimore2019bandit}.

Recently many studies focused on problems inspired from practical applications, in which additional prior information is available. For example the means of the arms might be known to be sparse  \citep{kwon2017sparse}, to form a Lipschitz function \cite{magureanu2014lipschitz} or have a linear structure \cite{dani2008stochastic,abbasi2011improved}, or a combinatorial one \cite{cesa2012combinatorial,kveton2015tight}. All these assumptions have been regrouped under the name ``structured bandit'' \cite{combes2017minimal}. As was the case in that last work, we aim at providing a family of algorithms that adapt to the known structure.

Bandit strategies realize a trade-off between exploitation (pulling the apparently best arm) and exploration. In the related problem of bandit pure exploration, in which no reward is gained but the goal is to answer a query, algorithms that adapt to any structure (and query) have been recently developed, first with asymptotic \cite{GK16}, then with non-asymptotic guarantees \cite{degenne2019non}. Our approach here will be to develop these techniques further, and obtain results for reward maximisation.

\subsection{Structured stochastic bandits}

A finite number $K$ of arms have reward distributions $(\nu_k)_{k\in [K]}$ in a known sub-Gaussian canonical exponential family with one parameter (the mean of the distribution) in an open interval $\Theta\subseteq \mathbb{R}$. The vector of means of these distributions is denoted by $\vmu\in\R^K$. The arm with highest mean for vector $\vmu$ is denoted by $i^*(\vmu)$ and we suppose that it is unique. We write $\mu^*$ for the value of that highest mean. An arm with mean below $\mu^*$ is called sub-optimal. For $x,y \in \Theta$, we denote by $d(x,y)$ the Kullback-Leibler divergence from the distribution parametrised by $x$ to that parametrised by $y$.

The mean vector $\vmu$ is known to belong to a set $\mathcal{M}\subseteq \Theta^K$, which represents the structure of the problem: the structural knowledge restricts the set of admissible mean parameters. For example, Lipschitz bandits prescribe that the means of successive arms cannot be far apart. We make the assumption that there exists a compact set $\mathcal C \subseteq \Theta^K$ such that $\mathcal M \subseteq \mathcal C$, which is convenient for the proof, but most likely not necessary. We also restrict our attention to what we call \emph{regular} structures: those verifying Assumption~\ref{ass:neg_is_lip} in appendix~\ref{app:regret_proof}. These include all examples of structures we found in the literature. The simplified Assumption~\ref{ass:regularity} below presents the same idea, is verified by all structures we consider but the Sparse one and is presented here instead of the more detailed one to preserve the brevity of the exposition.
For $j\in[K]$ we define $\neg j = \text{cl}(\setc*{\vmu {\in}  \mathcal{M}}{i^*(\vmu){\neq} j})$ the closure of the set of alternative structured problems where the best arm is not $j$. Let also $\neg_x i = \{\vlambda \in \mathbb{R}^K \ :\ \lambda^i = x \} \cap \neg i$.
\begin{assumption}[Regularity, simplified]\label{ass:regularity}
There exists $c_{\mathcal M}$ such that for all $\vmu \in \mathcal M$, denoting $i^*(\vmu) = *$, if $\neg_{\mu^*} * \neq \emptyset$ then for all $\vlambda \in \neg *$ there exists $\vxi \in \neg_{\mu^*} *$ such that for all $k\in[K]$, $\vert \xi^k - \lambda^k \vert \le c_{\mathcal M} \vert \mu^* - \lambda^* \vert$ .
\end{assumption}
%
The interaction between the algorithm and its environment is the following: at stage $t\ge 1$,
  \textbf{(1)} the algorithm pulls arm $k_t \in [K]$,
  \textbf{(2)} it observes a reward sample $X_t^{k_t} \sim \nu_{k_t}$,
  \textbf{(3)} its total reward is accrued by the mean reward $\mu^{k_t}$ (which is unobserved).
An algorithm is a sequence of functions, one for each time $t\in \mathbb{N}$, that take $(k_1, X_1^{k_1}, \ldots, k_{t-1}, X_{t-1}^{k_{t-1}})$ as input and return $k_t\in[K]$.

 We define the gap of an arm $k\in[K]$ as $\Delta^k = \mu^* - \mu^k$. Let $N_T^k$ be the number of pulls of arm $k$ up to time $T$. The goal of a bandit algorithm is to maximise its cumulated expected reward. Subtracting obtained reward from achievable reward, we arrive at the standard evaluation metric of expected regret,
\begin{align*}
\ex R_T
= T\mu^* - \sum_{t=1}^T \ex\mu^{k_t}
= \ex\sum_{t=1}^T \Delta^{k_t}
= \sum_{k=1}^K \Delta^k \ex N_T^k \: .
\end{align*}

\subsection{Contributions}

An algorithm is said to be asymptotically optimal if its regret verifies that for all $\vmu\in \mathcal M$, $\mathbb{E}_\vmu[R_t]/\ln(t)$ converges to a constant $V^{\mathcal M}(\vmu)$, which matches the constant prescribed by the corresponding lower bound (see Section~\ref{sec:lower_bound}).
\begin{itemize}[nosep]
  \item We introduce a family of algorithms that are asymptotically optimal for all structures $\mathcal M$, while having explicit non-asymptotic regret bounds.
  \item We exhibits members of that family with computational complexity much lower than that of earlier structure-adaptive algorithms for many structures, since they never solve fully the lower bound minimax problem.
  \item On experiments, we verify that the proposed algorithms adapt to the structure. Their regret is close to that of the UCB algorithm up to a time depending on the complexity $V^{\mathcal M}(\vmu)$ of the problem instance, after which the structural information is successfully exploited and the regret is of order $V^{\mathcal M}(\vmu)\ln T$. 
\end{itemize}

\subsection{Related work on structure adaptive methods}

Lower bounds for the regret of asymptotically consistent algorithms (i.e.\ with regret $o(T)$ for all $\vmu \in \mathcal M$) take the form of a constrained minimisation problem \cite{lai1985asymptotically,GravesLai97}.
The OSSB algorithm \cite{combes2017minimal} uses a test to decide whether to exploit or explore (we will use a similar mechanism).
If it explores, it solves a plug-in estimate of the lower bound problem, and pulls arms in order to approach the corresponding optimal sampling behaviour.
Its analysis is based on the continuity of the optimiser, seen as a function of the bandit instance $\vmu$: if the plug-in estimate is close enough to the true mean vector, the estimated pulling proportions are close to optimal.
In order to make sure that the empirical means converge to the true means, it pulls all arms $o(\ln T)$ times, a technique that is called ``forced exploration''.
The OSSB algorithm is claimed to be asymptotically optimal up to a multiplicative factor, which can be as close to $1$ as wanted, with the drawback that the asymptotic regime is delayed. 

A similar idea of tracking a plug-in estimate of the optimal pulling proportions was used before in the setting of fixed-confidence pure exploration, with the Track-and-Stop algorithm of \citet{GK16} (generalised to combinatorial settings by \citealt{pmlr-v65-chen17a}). Track-and-Stop achieves asymptotic optimality for any structure under the above continuity assumption \cite{mixmart}, and without it \citep{multiple.answers}. In recent work, a game-based algorithm was developed which gets non-asymptotic guarantees for any structure in the pure exploration setting \cite{degenne2019non}. We take inspiration from that line of work.

Our algorithm uses an explore or exploit test as is done in OSSB. If it explores, it treats the lower bound as the value of a zero-sum game between two players. We implement two regret-minimising algorithms, one for each player, and their interaction ensures convergence to the value of the game.

\section{Algorithm and Results}\label{sec:main}
Our goal is to design efficient algorithms for the problem of reward maximisation in structured stochastic multi-armed bandit models. We specifically aim to incur little regret. To calibrate our expectations about what can be achieved (at least in principle), we start by reviewing the classic asymptotic lower bound of \citep{GravesLai97}.

\subsection{Asymptotic Lower Bound}\label{sec:lower_bound}

Consider any algorithm, and assume that it is reasonable in the sense of \emph{asymptotic consistency}, meaning that for any $\vmu \in \mathcal M$, any sub-optimal arm $k \notin \argmax_i \mu^i$ is sampled only sub-polynomially often, i.e.\ $\ex_\vmu [N_T^k] \in \cap_{a > 0} o(T^\alpha)$.
\begin{theorem}[\citealt{GravesLai97}]
  An algorithm asymptotically consistent for bandit structure $\mathcal M$ must incur regret
  \[
    \liminf_{T \to \infty} \frac{\ex_\vmu \sbr*{R_T}}{\ln T} ~\ge~ V^{\mathcal M}(\vmu)
    \quad
    \text{for any instance $\vmu \in \mathcal M$}
  \]
  where the instance-dependent asymptotic regret rate $V^{\mathcal M}(\vmu)$ is the value of the optimisation problem (here $\Lambda \df \neg i^*(\vmu)$)
  \begin{equation}\label{eq:asymptotic.rate}
    \min_{\vN \ge \zeros} \sum_k N^k \Delta^k
    \quad
    \text{s.t.}
    ~
    \inf_{\vlambda \in \Lambda} \sum_k N^k d(\mu^k,\lambda^k) \ge 1
    .
  \end{equation}
\end{theorem}
Given this insurmountable limit, we take as our goal to construct algorithms that satisfy $\ex_\vmu[R_T] \le V^{\mathcal M}(\vmu) \ln T + o(\ln T)$, with explicit finite-time control over the lower-order term.

\subsection{Perturbed Game and Saddle Point Problems}
In a multi-armed bandit, the optimal arm has zero gap, $\Delta^* = 0$. This creates several technical complications later on, which we choose to avoid by picking a small $\varepsilon > 0$ and defining the \emph{$\varepsilon$-perturbed gaps} $\Delta_\varepsilon^k = \Delta^k \lub \varepsilon$. We call $V_\varepsilon^{\mathcal M}(\vmu)$ the instance-dependent regret rate \eqref{eq:asymptotic.rate} with these perturbed gaps substituted. We find three useful saddle point expressions.
\begin{lemma}
  The reciprocal $D_\varepsilon^{\mathcal M}(\vmu) \df 1/V_\varepsilon^{\mathcal M}(\vmu)$ satisfies
  \begin{subequations}
    \label{eq:perturbed.saddle.points}
\begin{align}
  \label{eq:pull.rates}
  D_\varepsilon^{\mathcal M}(\vmu)
  &~=~
  \max_{\w \in \triangle} \inf_{\vlambda \in \Lambda}~ \frac{\sum_k w^k d(\mu^k,\lambda^k)}{\sum_k w^k \Delta_\varepsilon^k}
  \\
  \label{eq:regret.rates}
  &~=~
  \max_{\tilde \w \in \triangle} \inf_{\vlambda \in \Lambda}~ \sum_k \tilde w^k \frac{d(\mu^k,\lambda^k)}{\Delta_\varepsilon^k}
  \\
  \label{eq:dual}
  &~=~
  \inf_{\q \in \triangle(\Lambda)}
  \max_k~ \frac{\ex_{\vlambda \sim \q}\sbr*{d(\mu^k,\lambda^k)}}{\Delta_\varepsilon^k}
\end{align}
\end{subequations}
\end{lemma}
Compared to \eqref{eq:asymptotic.rate}, problem \eqref{eq:pull.rates} is parametrised by the \textbf{fraction of rounds} $w^k \propto N^k$ spent pulling each arm $k$. The objective here is quasi-concave in $\w$ but not concave. The rewrite \eqref{eq:regret.rates} uses the \textbf{fraction of regret} $\tilde w^k \propto N^k \Delta_\varepsilon^k$ incurred by pulling arm $k$. This objective is linear (hence concave) in $\tilde \w$. From either form, randomising $\vlambda \sim \q$ licenses the min-max swap resulting in \eqref{eq:dual}.

These expressions correspond to a zero-sum game in which a pure strategy for the learner selects an arm $k \in [K]$, while for the opponent it picks a confusing instance $\vlambda \in \neg i^*(\vmu)$. The resulting payoff is then the information-regret ratio $\frac{d(\mu^k, \lambda^k)}{\Delta_\varepsilon^k}$, which quantifies the rate of progress in satisfying the information constraint in \eqref{eq:asymptotic.rate} per unit of objective value (i.e.\ regret) invested. In \eqref{eq:perturbed.saddle.points} the moves are ordered, upon which the outermost player needs to employ randomisation.

We quantify the cost of perturbation (Proof in Appendix~\ref{app:perturbation.cost})
\begin{theorem}\label{thm:perturbation.cost}
  Under Assumption~\ref{ass:regularity}, there is a $c > 0$ such that for small $\varepsilon$ we have $D_\varepsilon^{\mathcal M} \ge D^{\mathcal M} - c \sqrt{\varepsilon D^{\mathcal M}}$ and hence $V_\varepsilon^{\mathcal M} \le V^{\mathcal M} + c \sqrt{\varepsilon V^{\mathcal M}}$.
\end{theorem}

\subsection{Noise-Free Case}\label{sec:noisefree}
In this section we assume that we know the bandit model $\vmu \in \mathcal M$, and our goal is to compute the perturbed lower-bound value $V_\varepsilon^{\mathcal M}(\vmu)$ and a matching strategy $\vN$ from \eqref{eq:asymptotic.rate} or $\w$, $\tilde \w$ or $\q$ from \eqref{eq:perturbed.saddle.points}. Our approach will be to pick either form \eqref{eq:regret.rates} or \eqref{eq:dual}, and run an online learner for the outer player against best response for the inner player. We will run this interaction for a carefully selected number of rounds $n$ to get our result.

In the following we will call the maximising player, controlling $k$, the $k$-player, even if this player randomises $k \sim \tilde \w$. Similarly we will talk about the minimising $\vlambda$-player. To treat the structure in a modular way, we will make the following computational assumption:
\begin{assumption}\label{ass:altmin}
We are given an \emph{alt-min} oracle computing
\begin{equation}\label{eq:altmin}
  (\vmu, \w, j, k)
  ~\mapsto~
  \argmin_{\vlambda \in \mathcal M : \lambda_k \ge \lambda j}
  \sum_i w^i d(\mu^i, \lambda^i)
\end{equation}
for any vector $\vmu \in \Theta^K$, non-negative weights $\w$ and arms $j \neq k$. This implies tractability of argmin over $\mathcal M$ and $\neg j$.
\end{assumption}
This assumption is satisfied (either by a closed-form expression, a binary search or full-blown numerical convex optimisation) for all structures we use for the experiments in Section~\ref{sec:experiments} (unconstrained, sparse, linear, concave, unimodal and categorised).

\subsubsection{$\vlambda$-learner}
In this section we will target the saddle point \eqref{eq:dual} using an online learner for $\vlambda$. Each round $t$, this $\vlambda$-learner outputs a distribution $\q_t$ on $\Lambda$. Given $\q_t$, we define the $k$-opponent to play best response, i.e.\
\begin{equation}\label{eq:best.response.k}
  k_t
  ~\in~
  \argmin_k~ \frac{\ex_{\vlambda \sim \q_t}\sbr*{d(\mu^k, \lambda^k)}}{\Delta_\varepsilon^k}
  .
\end{equation}
We then have the $\vlambda$-learner update based on the linear loss function $\ell_t(\q) \df \ex_{\vlambda \sim \q}\sbr*{d(\mu^{k_t},\lambda^{k_t})}$. A $\vlambda$-learner regret bound of $\mathcal B_n^\vlambda$ for $n$ rounds of interaction gives us
\begin{align}\notag
  \sum_{t=1}^n \ell_t(\q_t)
  &~\le~
  \inf_{\vlambda \in \Lambda}~
  \sum_{t=1}^n d(\mu^{k_t}, \lambda^{k_t})
  + \mathcal B_n^\vlambda
  \\
  \label{eq:lambda.rbd}
  &~=~
  \inf_{\vlambda \in \Lambda}~
  \sum_k N_n^k d(\mu^{k}, \lambda^{k})
  + \mathcal B_n^\vlambda
  .
\end{align}
By definition of $k_t$,
\begin{align}
  \notag
  \sum_{t=1}^n \ell_t(\q_t)
  &~=~
    \sum_{t=1}^n \Delta_\varepsilon^{k_t}
    \frac{
    \ex_{\vlambda \sim \q_t}\sbr*{d(\mu^{k_t}, \lambda^{k_t})}
    }{
    \Delta_\varepsilon^{k_t}
  }
  \\
  \notag
  &~\stackrel{\text{\tiny \eqref{eq:best.response.k}}}{=}~
    \sum_{t=1}^n \Delta_\varepsilon^{k_t} \max_k
    \frac{
    \ex_{\vlambda \sim \q_t}\sbr*{
    d(\mu^{k}, \lambda^{k})
    }
    }{
    \Delta_\varepsilon^{k}
    }
    \:.
\end{align}
We then have, abbreviating $R_n^\varepsilon = \sum_{t=1}^n \Delta_\varepsilon^{k_t}$,
\begin{align}
  \notag
  \sum_{t=1}^n \ell_t(\q_t)
  &~\ge~
    \max_k \sum_{t=1}^n \Delta_\varepsilon^{k_t}
    \frac{
    \ex_{\vlambda \sim \q_t}\sbr*{d(\mu^{k}, \lambda^{k})}
    }{
    \Delta_\varepsilon^{k}
  }
  \\
  \notag
  &~=~
    R_n^\varepsilon \max_k \sum_{t=1}^n \frac{\Delta_\varepsilon^{k_t}}{R_n^\varepsilon}
    \frac{
    \ex_{\vlambda \sim \q_t}\sbr*{
    d(\mu^{k}, \lambda^{k})
    }
    }{
    \Delta_\varepsilon^{k}
  }
  \\
  \notag
  &~\ge~
    R_n^\varepsilon \inf_\q \max_k
    \frac{
    \ex_{\vlambda \sim \q}\sbr*{
    d(\mu^{k}, \lambda^{k})
    }
    }{
    \Delta_\varepsilon^{k}
  }
  \\
  \label{eq:lambda.folded}
  &~\stackrel{\text{\tiny \eqref{eq:dual}}}{=}~
  R_n^\varepsilon D_\varepsilon^{\mathcal M}(\vmu)
  .
\end{align}
We will choose $n$ adaptively, running the algorithm as long as $\inf_{\vlambda \in \Lambda} \sum_k N_n^k d(\mu^{k}, \lambda^{k}) \le \ln T$. Chaining \eqref{eq:lambda.rbd} and \eqref{eq:lambda.folded} then yield $R_n^\varepsilon D_\varepsilon^{\mathcal M}(\vmu) \le \ln T + \mathcal B_n^\vlambda$, so that
\[
  R_n ~\df~
  \sum_{t=1}^n \Delta^{k_t}
  ~\le~
  R_n^\varepsilon
  ~\le~
  V_\varepsilon^{\mathcal M}(\vmu) \del*{
    \ln T +
    \mathcal B_n^\vlambda
  }
  .
\]
At this point we need $\mathcal B_n^\vlambda$ to be $o(\ln T)$ for $V_\varepsilon^{\mathcal M}(\vmu)$ to be the dominant term. This will be feasible, as $n = \bigO(\ln T)$ and $\mathcal B_n^\vlambda$ can be taken to be $\bigO(\sqrt{n})$.

In particular, because $n \varepsilon \le R_n^\varepsilon \le V_\varepsilon^{\mathcal M}(\vmu) \del*{\ln T + c \sqrt{n}}$,
\begin{equation}\label{eq:explobdd}
  \sqrt{n} ~\le~ \frac{c V_\varepsilon^{\mathcal M}(\vmu) + \sqrt{c^2 V_\varepsilon^{\mathcal M}(\vmu)^2 + 4 \varepsilon V_\varepsilon^{\mathcal M}(\vmu) \ln T}}{2 \varepsilon}.
\end{equation}
So that indeed $\sqrt{n} = o(\ln T)$. Moreover, these asymptotics start kicking in once
$
  \ln T \gg
  \frac{c^2 V_\varepsilon^{\mathcal M}(\vmu)}{4 \varepsilon}
$. Conversely, to ensure asymptotic optimality we need to pick $\varepsilon = \omega(\wfrac{1}{\ln T})$. In our next section we will develop an anytime version using a time-decaying $\varepsilon_t$.

The practicality of implementing a $\vlambda$-learner depends on the geometry of the sets $\neg j$ and the makeup of the loss function. One may always write $\neg j$ as a union of $K-1$ cells, which are intersections of $\mathcal M$ with one linear inequality
\[
  \neg j
  ~=~
  \bigcup_{k \neq j}
  \setc*{\vlambda \in \mathcal M}{\lambda_k \ge \lambda_j}
  .
\]
For the structures considered in the experiments section (except the sparse structure), we find that these cells are in fact convex sets (while $\neg j$ is not). Moreover, for Gaussian and Bernoulli bandits the relative entropy function $d(\cdot,\cdot)$ is strongly convex in its second argument. Hence we can instantiate a sub-learner on each cell. An interesting choice is Follow-the-Leader, which is tractable by Assumption~\ref{ass:altmin} and incurs $\bigO(\ln t)$ regret. On the meta-level we aggregate the sub-learner iterates using an experts algorithm (we use AdaHedge by \citet{de2014follow} for $\bigO(\sqrt{t})$ regret).

\subsubsection{$k$-learner}
In this section we will target the saddle point \eqref{eq:regret.rates}. We will work with a $k$-learner that in each round $t$ outputs an action $\tilde \w_t$ (recall these are the desired per-arm fractions of regret, not of rounds). Given $\tilde \w_t$ from the $k$-learner, we define time proportions $\w_t$ and the best-response opponent $\vlambda_t \in \Lambda$ by
\begin{align}
  \label{eq:inverse.tilde.map}
  w_t^k &~\propto~ \tilde w_t^k /\Delta_\varepsilon^k,
  \\
  \label{eq:best.response.lambda}
  \vlambda_t &~\in~
  \argmin_{\vlambda \in \Lambda}~
  \sum_k w_t^k d(\mu^k, \lambda^k).
\end{align}
The $k$-learner then updates using the linear gain function
\begin{equation}\label{eq:gainfn}
  g_t(\tilde \w) ~=~ \del*{\sum_k w_t^k \Delta_\varepsilon^k} \sum_k \tilde w^k \frac{d(\mu^k, \lambda_t^k)}{\Delta_\varepsilon^k}.
\end{equation}
Note that the $w_t^k$ in its leading factor do \emph{not} vary with the argument $\tilde \w$; they are computed from $\tilde \w_t$ using \eqref{eq:inverse.tilde.map}.

A $k$-learner regret bound of $\mathcal B_n^k$ provides
\begin{equation}\label{eq:k.regret.bound}
\sum_{t=1}^n g_t(\tilde \w_t)
~\ge~
\max_k
\sum_{t=1}^n
\del*{\sum_j w_t^j \Delta_\varepsilon^j} \frac{d(\mu^k, \lambda_t^k)}{\Delta_\varepsilon^k}
- \mathcal B_n^k
.
\end{equation}
Moreover, the total gain satisfies
\begin{align}
  \notag
  \sum_{t=1}^n g_t(\tilde \w_t)
  &~=~
    \sum_{t=1}^n
    \del*{\sum_k w_t^k \Delta_\varepsilon^k} \sum_k \tilde w_t^k \frac{d(\mu^k, \lambda_t^k)}{\Delta_\varepsilon^k}
    \\
    \notag
    &~\stackrel{\text{\tiny \eqref{eq:inverse.tilde.map}}}{=}~
    \sum_{t=1}^n
    \sum_k w_t^k d(\mu^k, \lambda_t^k)
    \\
    \notag
    &~\stackrel{\text{\tiny \eqref{eq:best.response.lambda}}}{=}~
    \sum_{t=1}^n
    \inf_{\vlambda \in \Lambda}
    \sum_k w_t^k d(\mu^k, \lambda^k)
  \\
  \label{eq:k.folded}
  &~\le~
    \inf_{\vlambda \in \Lambda}~
    \sum_{t=1}^n
    \sum_k w_t^k d(\mu^k, \lambda^k).
\end{align}
Running as long as $\inf_{\vlambda \in \Lambda}
    \sum_{t=1}^n
    \sum_k w_t^k d(\mu^k, \lambda^k) \le \ln T$ and chaining \eqref{eq:k.regret.bound} with \eqref{eq:k.folded} results in
\begin{align}
\notag
\ln T
&~\ge~
\max_k
\sum_{t=1}^n
\del*{\sum_j w_t^j \Delta_\varepsilon^j} \frac{d(\mu^k, \lambda_t^k)}{\Delta_\varepsilon^k}
- \mathcal B_n^k
\\ 
&~=~
R_n^\varepsilon
\max_k
\sum_{t=1}^n
\frac{
  \del*{\sum_j w_t^j \Delta_\varepsilon^j}}{R_n^\varepsilon} \frac{d(\mu^k, \lambda_t^k)}{\Delta_\varepsilon^k}
- \mathcal B_n^k
\label{eq:SPk.chain1}
\end{align}
where we abbreviated $R_n^\varepsilon = \sum_{t=1}^n \sum_k w_t^k \Delta_\varepsilon^k$. Minimizing over $\q\in\triangle(\Lambda)$,
\begin{align}
\notag
\ln T
  &~\ge~
R_n^\varepsilon
\inf_{\q \in \triangle(\Lambda)}
\max_k~
    \frac{
    \ex_{\vlambda \sim \q} \sbr*{
    d(\mu^k, \lambda^k)}}{\Delta_\varepsilon^k}
- \mathcal B_n^k
\\
\label{eq:SPk.chain2}
&~=~
R_n^\varepsilon D_\varepsilon^{\mathcal M}(\vmu) - \mathcal B_n^k,
\end{align}
All in all we showed
\[
  R_n
  ~\df~
  \sum_{t=1}^n \sum_k w_t^k \Delta^k
  ~\le~
  R_n^\varepsilon
  ~\le~
  V_\varepsilon^{\mathcal M}(\vmu) \del*{
    \ln T
    +
    \mathcal B_n^k
  }
\]
and again we need to set things up so that $\mathcal B_n^k = o(\ln T)$. Now this may work, as we will have $n = \bigO(\ln T)$ and $\mathcal B_n^k = \bigO(\sqrt{n})$ so that, in total, $\mathcal B_n^k = \bigO(\sqrt{\ln T})$.

Tuning $\varepsilon$ is a bit trickier than for \eqref{eq:explobdd}, since the range of the gain function \eqref{eq:gainfn} scales with $1/\varepsilon$ and so hence will $\mathcal B_n^k$. Still $n = \bigO(V_\varepsilon/\varepsilon \ln T)$ and hence the asymptotics kick in for $\ln T \gg \del{\wfrac{V_\varepsilon}{\varepsilon}}^3$, or equivalently $\varepsilon \gg \wfrac{V_\varepsilon}{(\ln T)^{1/3}}$. We expect that power of $T$ is artificial; the range of practically observed gains \eqref{eq:gainfn} may well be constant.

A bandit algorithm cannot pull any $\w_t$ in the simplex, as would be prescribed by the $k$-learner. It has to pull one arm at each time. We circumvent that difficulty by using a so-called tracking procedure, which ensures that for all times $\vN_t \approx \sum_{s=1}^t \w_s$. We will use $k_t \in \argmin_{k\in[K]} N_{t-1}^k - \sum_{s=1}^t w_s^k$ (breaking ties arbitrarily). A similar rule was analysed in \citet{GK16}. Our analysis reveals that it ensures that for all $t\in \mathbb{N}$ and $k\in [K]$,
\begin{align*}
-\ln(K) \le N_t^k - \sum_{s=1}^t w_s^k \le 1 \: .
\end{align*}
The previous result \citep[Lemma~15]{GK16} has $K-1$ instead of our $\ln K$. Our proof makes use of a subtle invariant; it can be found in Appendix~\ref{app:tracking}.

In terms of implementation, we need to supply two things. First, a learner for linear losses defined on the simplex. This is a standard experts problem, for which we use AdaHedge \citep{de2014follow}. Second, we need to compute the best response $\vlambda_t \in \neg i^*(\vmu)$. This is where the structure dependence of the approach manifests; this step is tractable by Assumption~\ref{ass:altmin}.
The computational complexity of either the $k$-learner or $\vlambda$-learner based iterative saddle point approach is dominated by $K-1$ alt-min oracle calls every round.

\subsection{Saddle Point-Based MAB Reward Maximisation}
In this section we build atop the noise-free saddle point computation presented above. We obtain algorithms for regret minimisation in structured bandit problems that have finite-time bounds which convey, in particular, asymptotic optimality. The k-learner variant is displayed as Algorithm~\ref{alg:SPk}.

The main challenge is that we do not know $\vmu$ (nor anything derived, including the gaps $\bm\Delta_\varepsilon$ and the index of the best arm $i^*(\vmu)$), and that we hence need to estimate these live.

\begin{algorithm}[t]
  \begin{algorithmic}[1]
    \REQUIRE Exploration threshold $f(t)$
    \REQUIRE Learner $\mathcal{A}$ for linear losses on the simplex
    \STATE Start an instance $\mathcal{A}_j$ for each arm $j$.
    \STATE Pull each arm once and get $\hat{\vmu}_K$.
    \STATE Initialise $N_t^k = 1$ and $n_t^k = 0$ for all $k \in [K]$.
\FOR{$t = K+1,\cdots,T$}
	\STATE \textbf{if} there is an exploit-worthy $i$ for which \eqref{eq:explore/exploit} \textbf{then}
	\STATE $k_t = i$ (pick any if there are several suitable).
	\STATE \textbf{else}
    \STATE Estimate best arm: $j_t = \argmax_k \hat \mu_{t-1}^k$.
    \STATE \textbf{if} $n_t^{j_t}$ is even \textbf{then} $k_t = j_t$. \textbf{else}
        \STATE Compute confidence interval $[\underline{\mu_{t-1}^k}, \overline{\mu_{t-1}^k}]$ for every arm $k\in[K]$  using \eqref{eq:small.CI}.
        \STATE Estimate gaps: $\tilde \Delta_{\varepsilon_t}^k = \max \set[\big]{\varepsilon_t, \overline{\mu_{t-1}^{j_t}} - \overline{\mu_{t-1}^k}}$.
	      \STATE Get $\tilde\w_t$ from $\mathcal{A}_{j_t}$, compute $w_t^k \propto \tilde w_t^k/\tilde \Delta_{\varepsilon_t}^k$.
        \STATE Find the best response in $\neg j_t$ to $\w_t$ given $\hat\vmu_{t-1}$:
        \[
          \vlambda_t
          ~=~
          \argmin_{\vlambda\in\neg j_t}\sum_k w_t^k d(\hat{\mu}^k_{t-1}, \lambda^k)
          .
          \]
        \STATE Compute estimates $\UCB_t^k$ as in \eqref{eq:UCBs}
        \STATE Feed $g_t(\tilde \w) {=} \del*{\sum_k w_t^k \tilde \Delta_{\varepsilon_t}^k} \sum_k \tilde w^k \UCB_t^k$ to $\mathcal{A}_{j_t}$.
	\STATE Pull $k_t = \argmin_{k\in [K]} N_{t-1}^k - \sum_{s=1}^t w_s^k$. 
  \STATE \textbf{end if}
  \STATE Increase $n_t^{j_t}$ by 1.
  \STATE \textbf{end if}
	\STATE Access reward $X_t^{k_t}$, update $\hat{\vmu}_t$ and $\vN_t$.
\ENDFOR
\end{algorithmic}
\caption{$\text{SP}_k$ Learner}\label{alg:SPk}
\end{algorithm}

We sketch an outline of the construction. The main distinction made by our algorithm is whether to \emph{explore} or \emph{exploit}. Exploitation occurs when we are certain enough that we have the right best arm for our bandit $\vmu$, that is when
\begin{equation}\label{eq:explore/exploit}
  \exists i \in[K], \: \min_{\vlambda\in\neg i}\sum_k N_{t-1}^k d(\hat{\mu}^k_{t-1}, \lambda^k)
  ~>~
  f(t-1)
  ,
\end{equation}
where we will use a threshold $f(t)$ ($\approx \ln t$ plus lower order) that is high enough so that the cumulative contribution to the regret of rounds where \eqref{eq:explore/exploit} fails is bounded by a constant. Note that we cannot afford a failure probability $ \ge \frac{1}{t}$, for then the contribution of the failure cases would be of order $\ge \ln T$, voiding asymptotic optimality.

During the other rounds, our algorithm explores. The main idea here is to pick an online learner for either the $k$ or $\vlambda$ side, and adapt the corresponding noise-free pipeline \eqref{eq:SPk.chain1} and \eqref{eq:SPk.chain2} or \eqref{eq:lambda.rbd} and \eqref{eq:lambda.folded}. These rounds all happen under the complement of \eqref{eq:explore/exploit}, which now fulfils the role of the stopping condition ``running as long as \dots'' in Section~\ref{sec:noisefree}.

We now go over the major upgrades one by one, relating back to our Algorithm~\ref{alg:SPk} as we go along.

\paragraph{Estimating $i^*(\vmu)$.} First, in Section~\ref{sec:noisefree} we assumed we know the best arm $i^*(\vmu)$. Here that is no longer the case, and instead we have to rely on estimates, which can be wrong. Our approach to deal with this is to run not one but $K$ many saddle point computations, one for every candidate best arm. In each exploration round, we produce an estimate $j_t$ of the best arm, and only advance the saddle point interaction corresponding to that estimate. This way, the saddle point iteration for arm $i^*(\vmu)$ will simulate a pure trajectory as per Section~\ref{sec:noisefree}. Using concentration of measure and invoking our learners' regret bounds, we show (in Appendix~\ref{sec:misplore}) that the other instances corresponding to incorrect estimates of the best arm will only make a lower-order $\bigO(\ln \ln T)$ contribution to the regret. With that out of the way, we can reason about exploration rounds in which $j_t = i^*(\vmu)$.

\paragraph{Estimating $\vmu$.} Our second problem is then that we need to estimate the bandit instance $\vmu$ and its perturbed sub-optimality gaps $\Delta_\varepsilon^k$. We will be guided by the following idea. We want to look at the noise-free chain \eqref{eq:SPk.chain1} and \eqref{eq:SPk.chain2}, and relate all occurrences of estimates $\hat \vmu$, $\tilde \Delta_{\varepsilon_t}^k$, to their unknown truth $\vmu$ and $\Delta_{\varepsilon_t}^k$. We will take inspiration from the classic UCB analysis, and use concentration to bound the instantaneous estimation error after $n$ exploration rounds by $\sqrt{\ln(n)/n}$. These errors then accumulate over $n$ exploration rounds to an order $\sqrt{n \ln n}$ overhead. This is a lower-order term as long as the number $n$ of exploration steps is $o(\ln T)^2$.

More precisely, let $E_{s}^*$ be the event that we explore at stage $s$ and have the right guess for $i^*(\vmu)$ and define $n_t^* = \sum_{s=1}^t \indicator\{E_s^*\}$. With estimates $\hat{\mu}_t$, we have similarly to equation~\eqref{eq:k.folded} that if we explore at state $t+1$,
\begin{align*}
\ln t
&\ge \inf_{\vlambda \in \Lambda } \sum_{s\le t: E_s^*}\sum_{k=1}^K w_s^k d(\hat{\mu}_t^k, \lambda^k).
\end{align*}
Our algorithm cannot play at time $s<t$ based on $\hat{\vmu}_t$, since it does not know it yet. It will use the available estimate $\hat{\vmu}_{s-1}$. We then quantify how far these estimated means are from $\vmu$ with a concentration event $\mathcal E_n^{exp}$, which states that for all $k\in [K]$ and all times $t$,
\begin{align}
  \notag
&\mu^k \in [\underline{\mu_t^k}(n), \overline{\mu_t^k}(n)], \text{ with }\underline{\mu_t^k}(n), \overline{\mu_t^k}(n) \text{ solution to}
  \\
  \label{eq:small.CI}
&N_t^k d(\hat{\mu}_t^k, x) = \ln(n) + \bigO(\ln\ln t) \: .
\end{align}
That event is such that apart from a small number of rounds, when we explore event $\mathcal E_{n_t^*}^{exp}$ happens. If we can ensure that $n_t^*$ is less than a power of $\ln(t)$, then these intervals are of order $\ln\ln t$. We get
\begin{align*}
\ln t
\ge \inf_{\vlambda \in \Lambda } \sum_{s\le t: E_s^*\cap \mathcal E_{n_t^*}^{exp}}\sum_{k=1}^K w_s^k d(\hat{\mu}_{s-1}^k, \lambda^k) - \bigO(\sqrt{n_t^*})
\: .
\end{align*}
In order to continue along computations~\eqref{eq:k.folded},~\eqref{eq:SPk.chain1} and~\eqref{eq:SPk.chain2}, we need estimates for the gaps. Based on the small confidence intervals $[\underline{\mu_t^k}, \overline{\mu_t^k}]$ (where $\overline{\mu_t^k}=\overline{\mu_t^k}(n_t^*)$), we define $\tilde{\Delta}_t^k = \max\{\varepsilon, \overline{\mu_t^{j_t}} - \overline{\mu_t^{k}}\}$. Then up to a $\bigO(\sqrt{n_t^*})$ term
\begin{align*}
\ln t
&\ge \inf_{\vlambda \in \Lambda } \sum_{s\le t: E_s^*\cap \mathcal E_{n_t^*}^{exp}} (\sum_{j=1}^K w_s^j \tilde{\Delta}_s^j)\sum_{k=1}^K \tilde{w}_s^k \frac{d(\hat{\mu}_{s-1}^k, \lambda^k)}{\tilde{\Delta}_s^k}
\\
&\ge \sum_{s\le t: E_s^*\cap \mathcal E_{n_t^*}^{exp}} (\sum_{j=1}^K w_s^j \tilde{\Delta}_s^j)\sum_{k=1}^K \tilde{w}_s^k \frac{d(\hat{\mu}_{s-1}^k, \lambda_s^k)}{\tilde{\Delta}_s^k}
\: .
\end{align*}
\paragraph{Optimism.} Using the estimated gains directly could lead to high regret, in the same way as the Follow-The-Leader strategy might accrue linear regret in stochastic bandits. We use the now widely employed Optimism in Face of Uncertainty principle \cite{agrawal1995sample} and replace the gains by an optimistic estimate.
In the hypothetical noiseless case, we would like at time $s$ to feed to the $k$-learner the gain vector $k \mapsto (\sum_{i=1}^K w_s^i \Delta^i_{\varepsilon_s})\frac{d(\mu^k, \lambda^k_s)}{\Delta^k_{\varepsilon_s}}$ (see \eqref{eq:gainfn}). We define
\begin{align}\label{eq:UCBs}
\UCB_t^k
&= \max_{\xi \in [\underline{\mu_{t-1}^k}, \overline{\mu_{t-1}^k}]} \frac{d(\xi, \lambda_s^k)}{\max\{\varepsilon_t, \indicator\{k\neq j_t\}(\overline{\mu_{t-1}^{j_t}} - \xi)\}}.
\end{align}
Under the event that $\mu^k\in [\underline{\mu_{t-1}^k}, \overline{\mu_{t-1}^k}]$, this is an upper bound for the factor depending on $k$ in the gain of the $k$-learner, up to the difference between $\overline{\mu_{t-1}^{j_t}}$ and $\mu^{j_t}$, which will be small since we ensure that $j_t$ is pulled every other round in the exploration phase. Again under that event, $\tilde{\Delta}_{t-1}^k(\UCB_t^k- d(\hat{\mu}_{t-1}^k, \lambda_s^k)/\tilde{\Delta}_{t-1}^k)$ is upper bounded, such that the total cost of the introduction of those upper confidence bounds is $\bigO(\sqrt{n_t^*})$.  We can now finish the computations of equations~\eqref{eq:SPk.chain1} and~\eqref{eq:SPk.chain2}: up to $\bigO(\sqrt{n_t^*})$,
\begin{align*}
\ln t
&\ge \sum_{s\le t: E_s^*\cap \mathcal E_{n_t^*}^{exp}} (\sum_{j=1}^K w_s^j \tilde{\Delta}_s^j)\sum_{k=1}^K \tilde{w}_s^k \UCB_s^k
\\
&\ge \max_{k\in K} \sum_{s\le t: E_s^*\cap \mathcal E_{n_t^*}^{exp}} (\sum_{j=1}^K w_s^j \tilde{\Delta}_s^j) \UCB_s^k
\\
&\ge \max_{k\in K} \sum_{s\le t: E_s^*\cap \mathcal E_{n_t^*}^{exp}} (\sum_{j=1}^K w_s^j \tilde{\Delta}_s^j) \frac{d(\mu^k, \lambda^k_s)}{\Delta^k_{\varepsilon_s}}
\\
&\ge R_t^{\varepsilon,*} D_\varepsilon^{\mathcal M}(\vmu) - \bigO(\sqrt{n_t^*})
\: .
\end{align*}
Finally, since $n_t^* \le R_t^{\varepsilon,*}/\varepsilon$ this is an equation which asymptotically gives $R_T^{\varepsilon,*} \le V_\varepsilon^{\mathcal M}(\vmu)\ln T$ as desired. Since this is the only term in the regret which is not $o(\ln T)$, we have asymptotic optimality if we have $\varepsilon \to 0$ at a suitable rate.

\paragraph{Effect of the confidence intervals on the sampling behaviour.} During the first rounds, the algorithm does not exploit yet. Hence the number of exploration rounds $n_t$ is $t$. If the best arm is pulled enough to be well estimated, our gap estimates are $\tilde{\Delta}_t^k \approx \mu^{j_t} - \mu^k - \sqrt{2\ln(n_t)/N_t^k}$ in the Gaussian case. For these estimates to be representative of the true gaps $\mu^{j_t}-\mu^k$, we need $N_t^k \propto \ln(n_t)/(\Delta^k)^2$. This is of order $\ln\ln t$ once $n_t$ is logarithmic in $t$, but it is $\ln(t)/(\Delta^k)^2$ at the beginning. We indeed verify experimentally that during the first rounds our algorithm pulls similarly to vanilla UCB. Afterwards, the exploitation regime starts and the structure adaptive sampling begins.

\paragraph{How our algorithm relates to previous techniques.}

Our algorithm employs a explore/exploit test based on a log-likelihood ratio, as is done in the OSSB algorithm \cite{combes2017minimal}. While the exploration phase of OSSB is akin to the exploration algorithm Track-and-Stop \cite{garivier2016optimal}, our exploration phase is inspired from the game point of view of \cite{degenne2019non}. The strategy presented above generalizes that exploration algorithm in several key aspects (the pure exploration case is recovered by forcing all gaps $\Delta^k$ to 1). 
\begin{itemize}[nosep]
	\item addressing the regret saddle-point problem with online learning algorithms is non-trivial even for known gaps, and requires several innovations (Section \ref{sec:noisefree}). For example, one of the players has to play regret proportions instead of pull proportions, as in~\eqref{eq:inverse.tilde.map}, and gap-weighted regret bounds need to be employed, as in~\eqref{eq:gainfn}.
	\item the unknown gaps need to be estimated appropriately (line 11 of Alg.~\ref{alg:SPk}),
	\item one of the gaps is 0, which is a problem since we divide by the gaps: we introduce the epsilon-perturbed problem to solve that issue,
	\item we need to relate the values of the perturbed and original games (Theorem~\ref{thm:perturbation.cost}).
\end{itemize}

\subsection{Main Result}

\begin{theorem}
For any $k$-learner or $\lambda$-learner with regret bound of order $\bigO(\sqrt{n})$ after $n$ steps, the expected regret of our structure adaptive algorithm verifies for all $\vmu \in \mathcal M$,
\begin{align*}
\lim_{T\to +\infty} \frac{\mathbb{E}_{\vmu}[R_T]}{\ln T}
\le V^{\mathcal M}(\vmu)
\: .
\end{align*}
\end{theorem}
The arguments presented above in fact lead to a finite time bound, but it contains many $o(\ln T)$ terms that we choose not to detail here. Proof in Appendix~\ref{app:regret_proof}.

\section{Experiments}\label{sec:experiments}

\begin{figure*}[t!]
  \centering
\subfigure[Unconstrained bandits {$\vmu = [0, 0.33, 0.67, 1]$}.]{\includegraphics[width=0.32\textwidth]{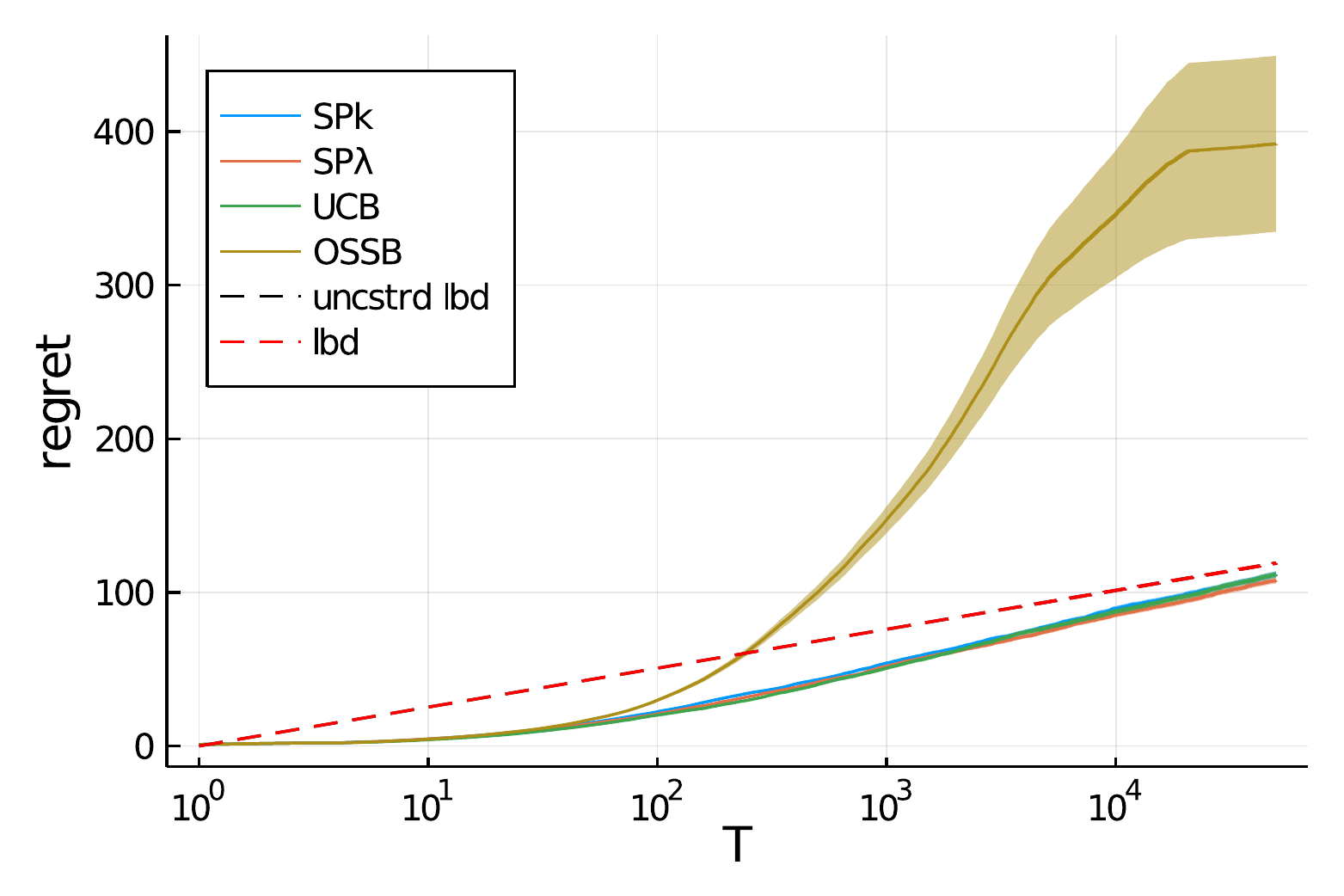}\label{fg:unconstrained1}}
\subfigure[Categorised bandits with {$\vmu^1=[2]$} and {$\vmu^2 = [1, 0.96, 0]$}.]{\includegraphics[width=0.32\textwidth]{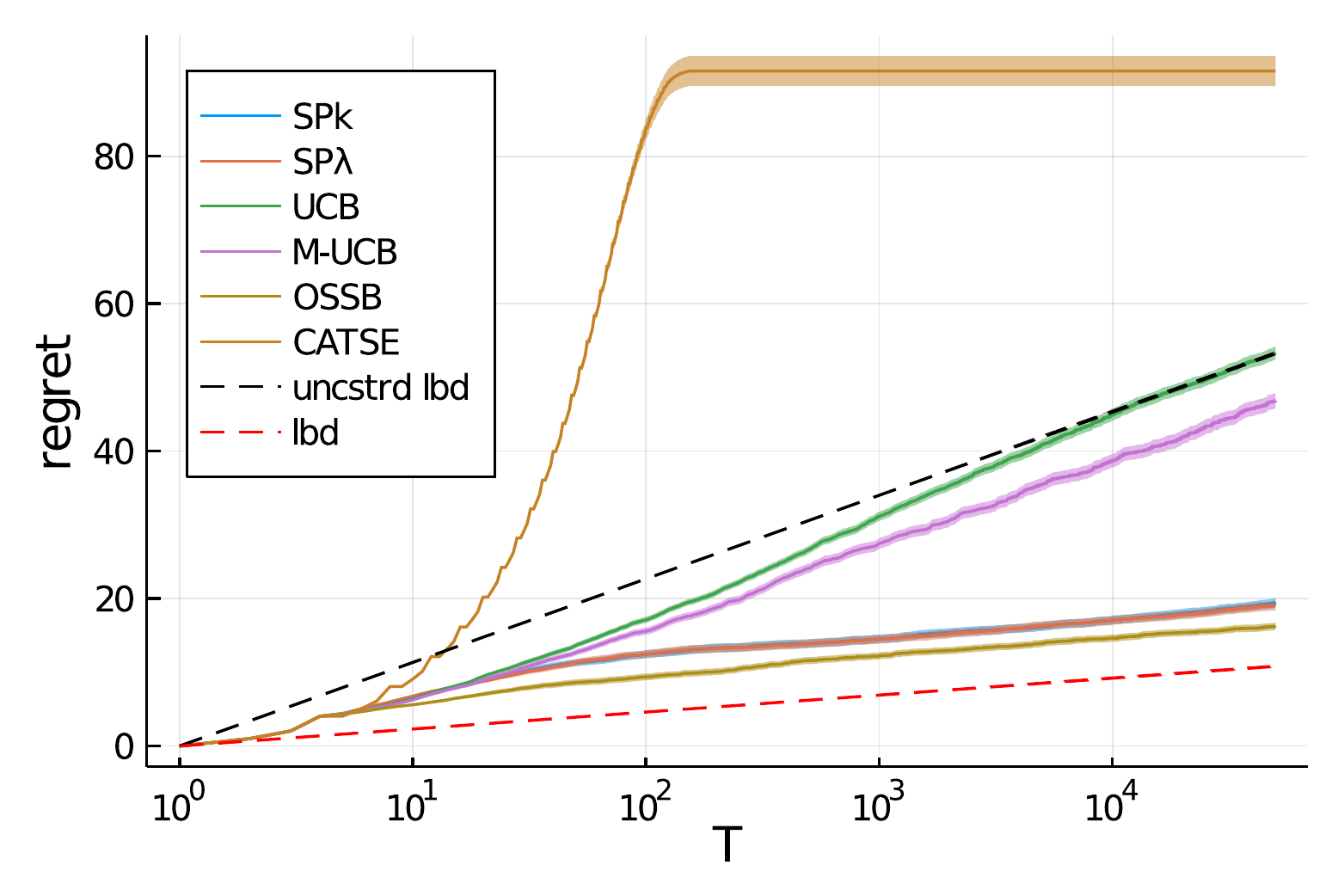}\label{fg:categorized1}}
\subfigure[Categorised bandits with {$\vmu^1=[3 ,1.50]$} and {$\vmu^2 = [1.44, 1.44, 0]$}.]{\includegraphics[width=0.32\textwidth]{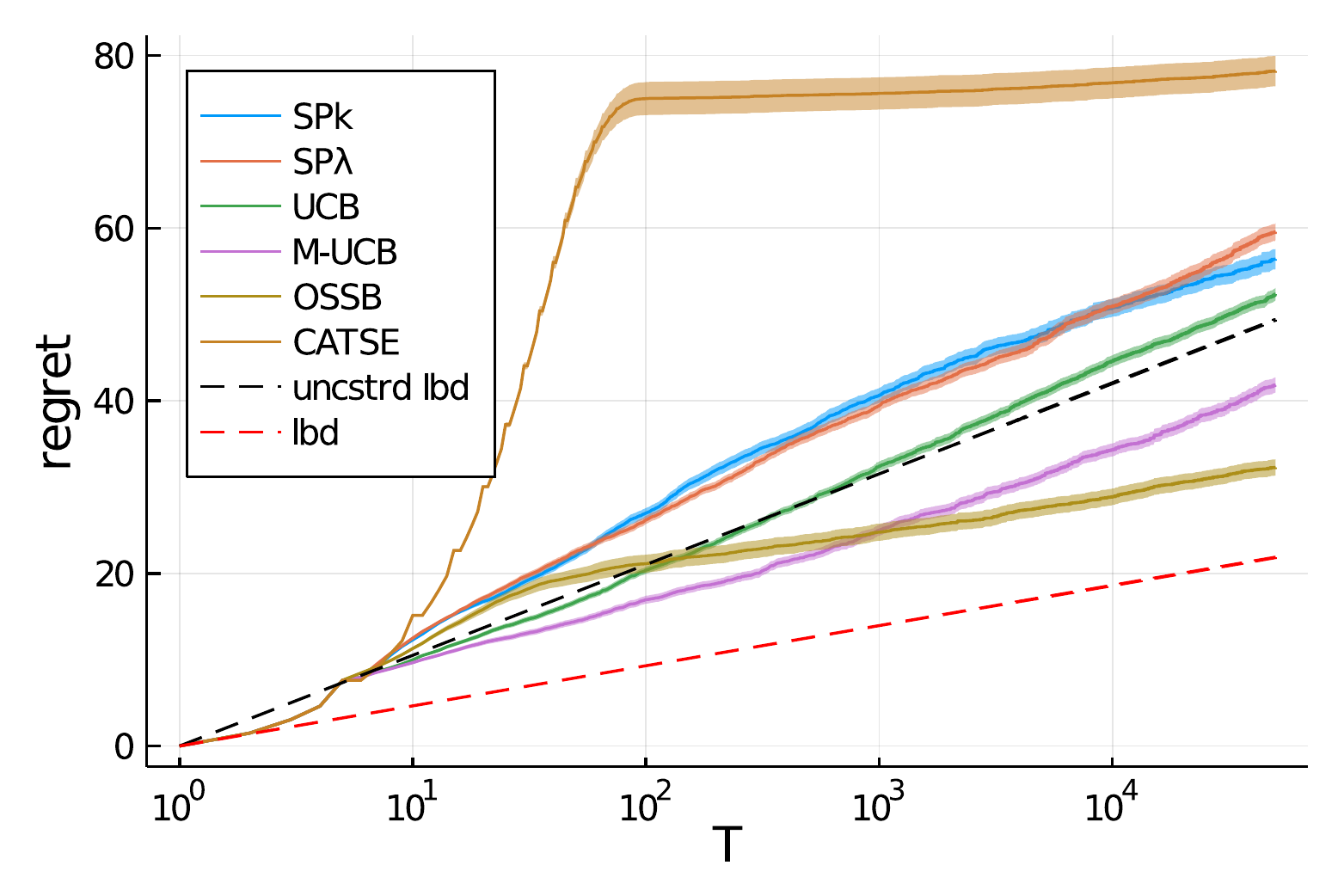}\label{fg:categorized2}}
\subfigure[Linear bandits with arms {$\x = [(\cos(\rho),\sin(\rho)), \rho \in \{2i\pi/5, 2i\pi/5+0.15\}_{i=0}^4]$} and $\theta = (0, 2)$.]{\includegraphics[width=0.32\textwidth]{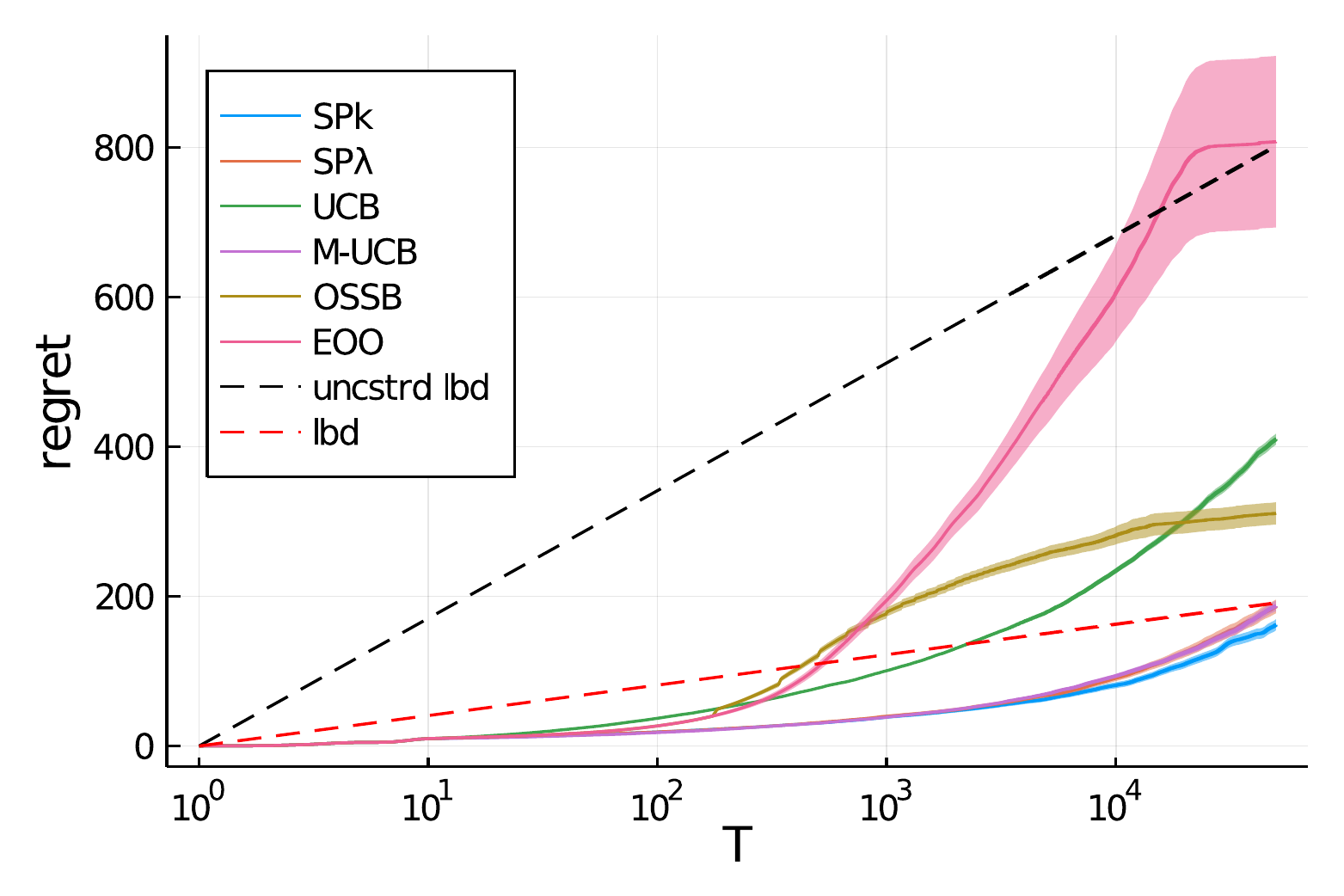}\label{fg:linear1}}
\subfigure[Unimodal bandits with {$\vmu = [0.2,0.4,0.9,0.7,0.1]$}.]{\includegraphics[width=0.32\textwidth]{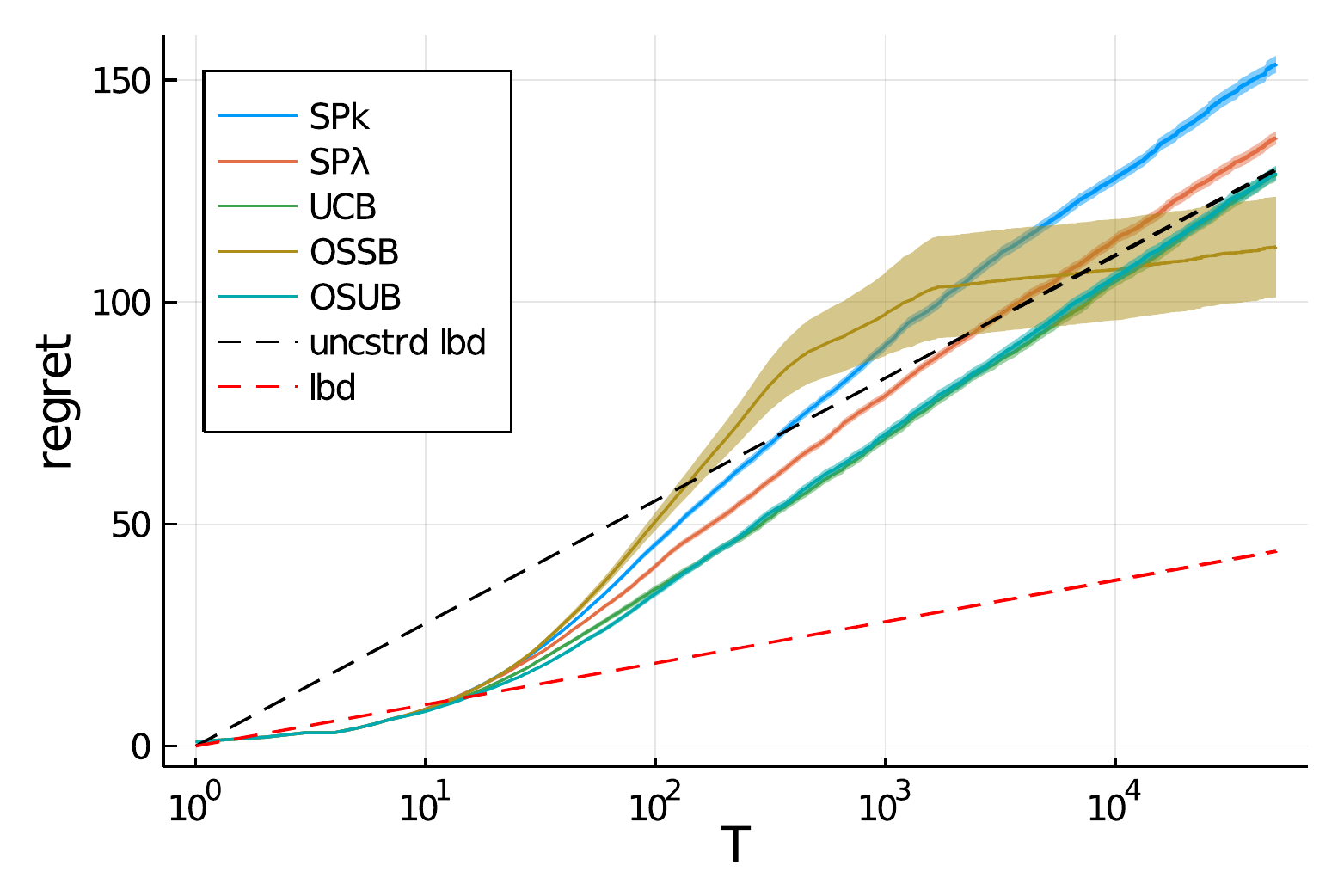}\label{fg:unimodal1}}
\subfigure[Unimodal bandits with {$\vmu = [0,0,0,1.00,0,0,0]$}.]{\includegraphics[width=0.32\textwidth]{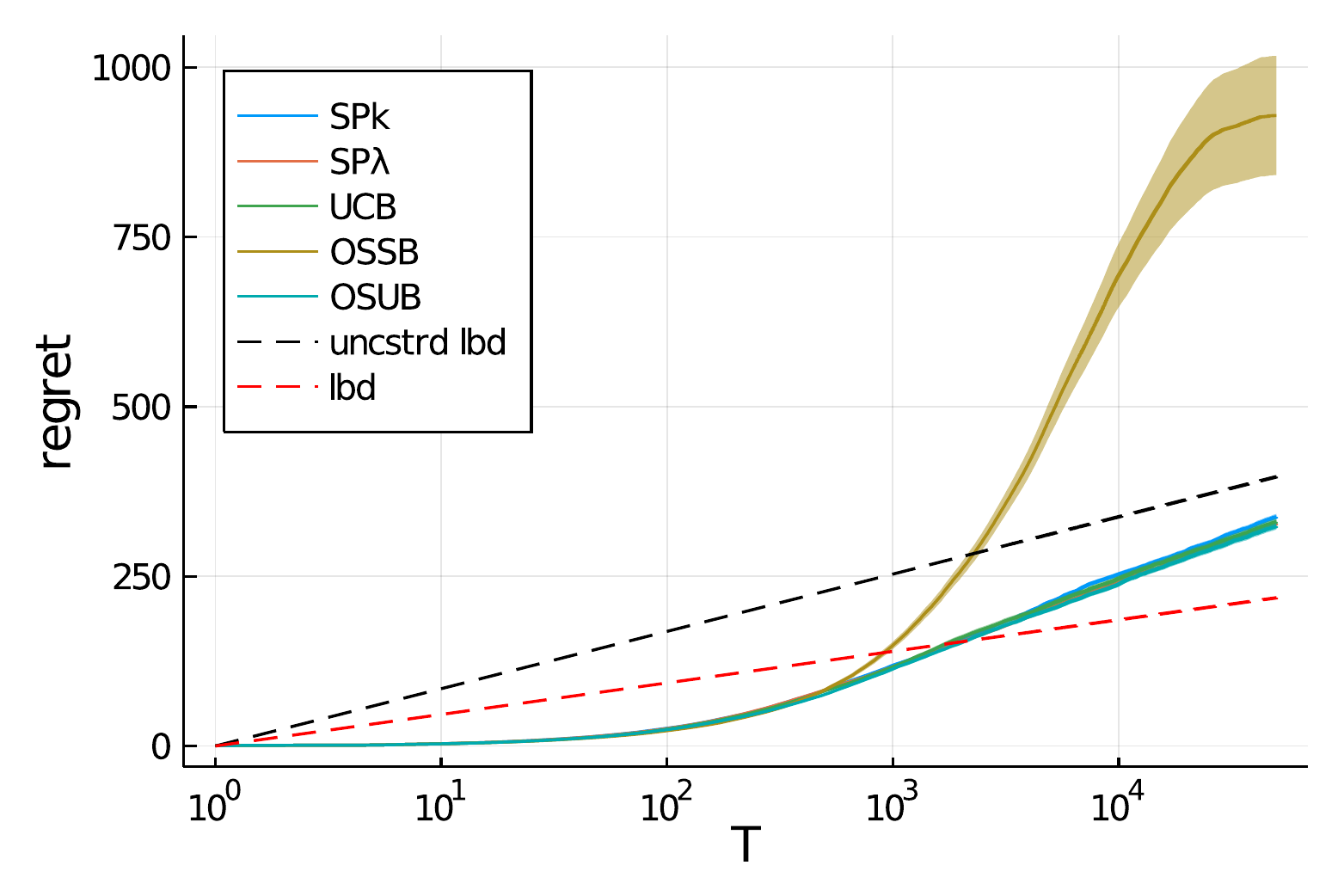}\label{fg:unimodal2}}
\subfigure[Sparse bandits with $s=1$, $\gamma = 0.3$ and $\mu_1 = 0.8$.]{\includegraphics[width=0.32\textwidth]{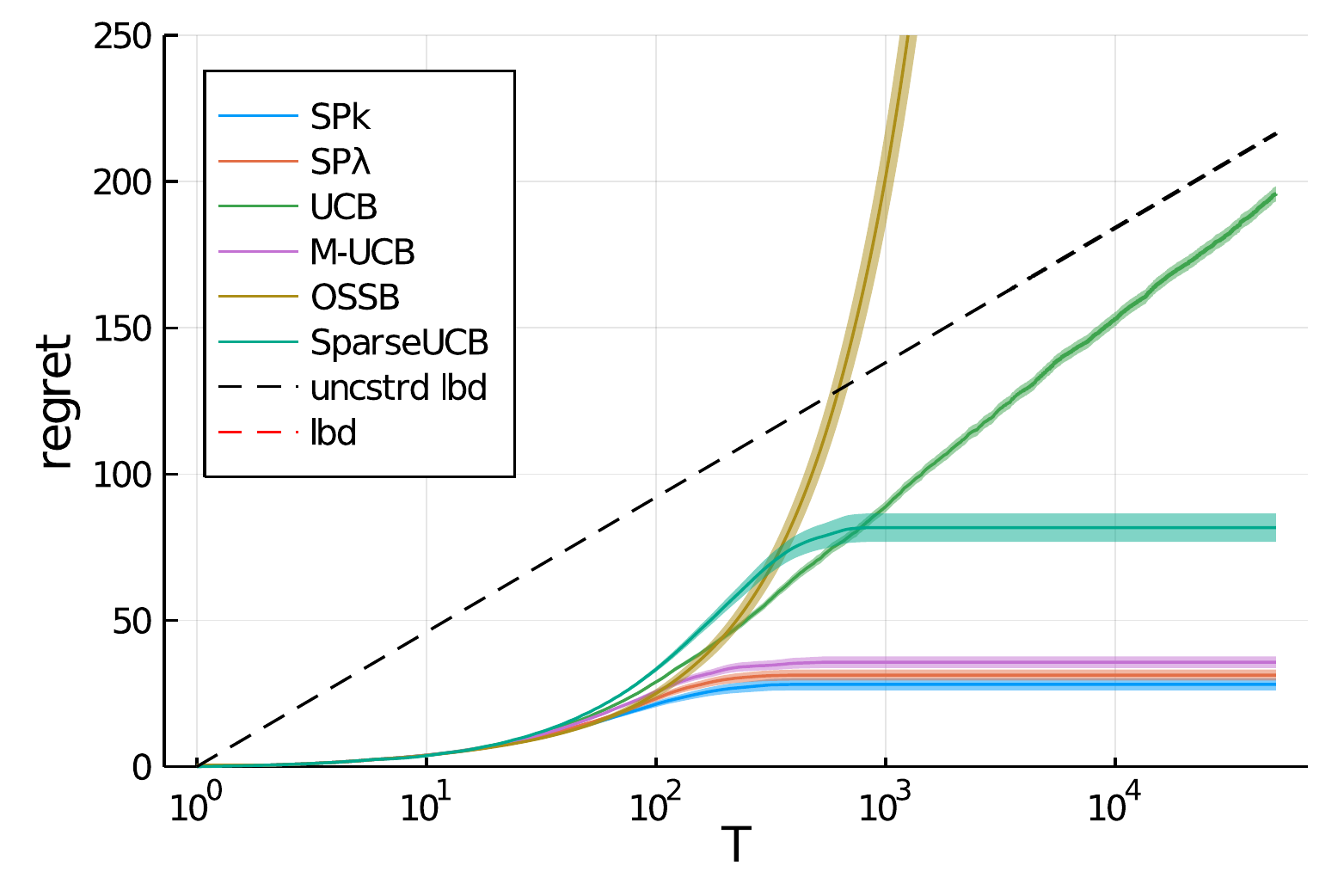}\label{fg:sparse1}}
\qquad
\subfigure[Sparse bandits with $s=2$, $\gamma = 0$ and ${[\mu_1,\mu_2] = [3.00,2.00]}$.]{\includegraphics[width=0.32\textwidth]{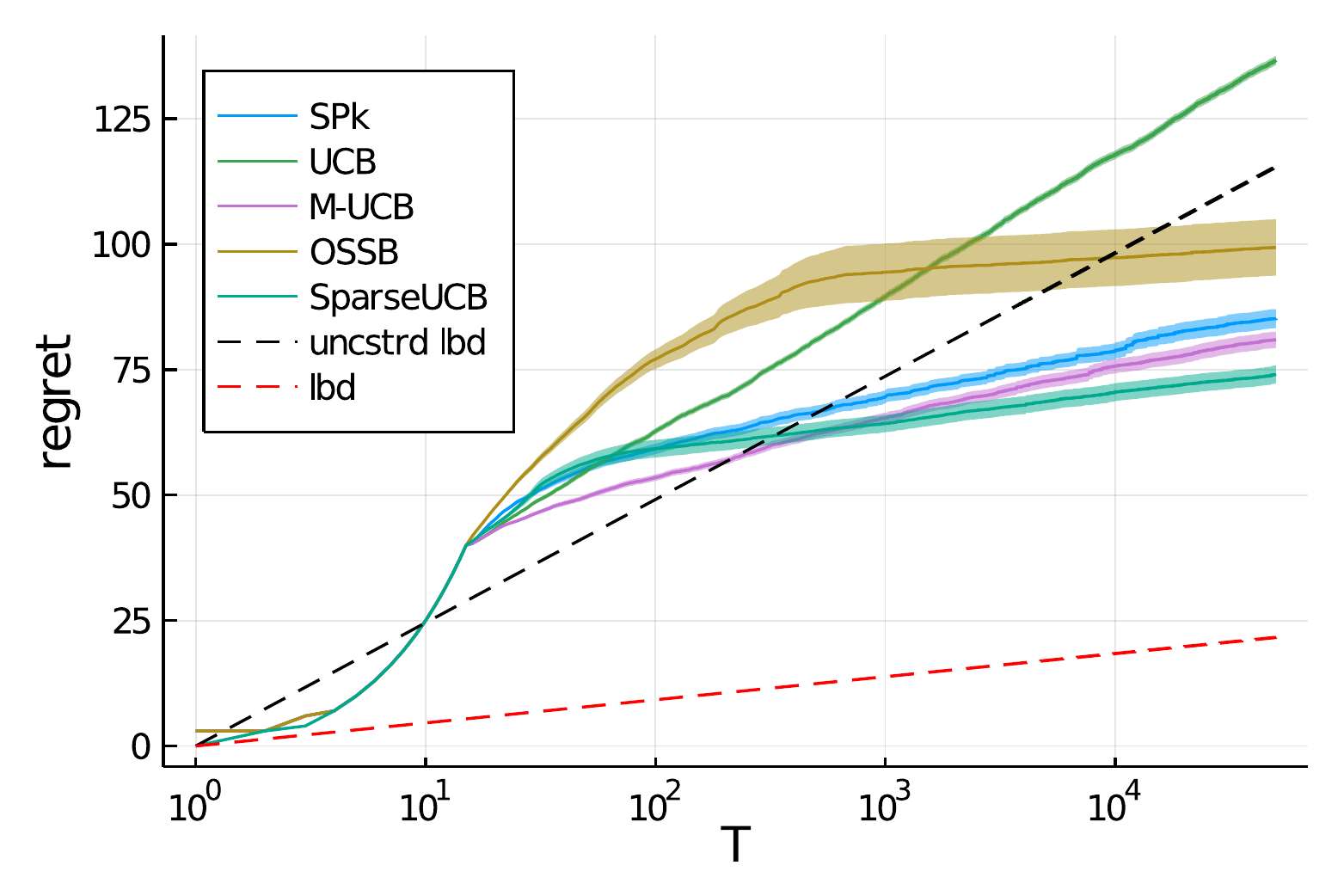}\label{fg:sparse2}}
  \caption{Regret of algorithms.}\label{fg:exp}
\end{figure*}

As mentioned in Section~\ref{sec:noisefree}, the asymptotics kick in when $\ln T \gg \frac{c^2 V_{\varepsilon_T}^{\mathcal M}(\vmu)}{4 \varepsilon_T}$ for the $\vlambda$-learner and $\ln T \gg \del{\wfrac{V_{\varepsilon_T}^{\mathcal M}(\vmu)}{\varepsilon_T}}^3$ for the $k$-learner.
This indicates that for small time horizons, the asymptotic theoretical guarantees become meaningless. It is worth noting that this is ubiquitous in the asymptotic literature. We still investigate empirically whether and when these algorithms start exploiting the structures and how they perform in small horizons. Implementation details are in Appendix~\ref{sec:impldet}.

We perform four series of experiments. In all cases, rewards are Gaussian with variance 1. The structures are:
\begin{itemize}[nosep]
  \item Figure~\ref{fg:unconstrained1}: unconstrained. This is the usual stochastic bandit setting, without additional structure.
  \item Figures~\ref{fg:categorized1} and~\ref{fg:categorized2}: categorised bandits with strong dominance \cite{jedor2019categorized}. The arms belong to one of two categories (and this information on each arm is known), and all arms of one category have means higher than all arms of the other (but which category is the ``good'' one is unknown).
  \item Figure~\ref{fg:linear1} 
  : linear \cite{auer2002using}. There exists an unknown parameter $\theta \in \mathbb{R}^d$ with $d<K$ and known vectors $x_1, \ldots, x_K \in \mathbb{R}^d$ such that $\mu^k = x_k^\intercal \theta$.
  \item Figures~\ref{fg:unimodal1} and~\ref{fg:unimodal2}: unimodal \cite{combes2014unimodal}. The mean function $k\mapsto \mu^k$ is unimodal, i.e. $\mu^1\leq \mu^2\leq \ldots\leq\mu^*\geq\ldots\geq \mu^{K-1}\geq \mu^{K}$.
  \item Figures~\ref{fg:sparse1} and~\ref{fg:sparse2}: sparse \cite{kwon2017sparse}. Given sparsity $s$ and value $\gamma$, $s$ of $K$ arms have means above $\gamma$ while others have means $\gamma$.
\end{itemize}

We mainly compare the following four algorithms,
\begin{itemize}[nosep]
\item SP$_k$, our saddle point algorithm based on a $k$-learner. The learner used is AdaHedge \cite{de2014follow}.
\item SP$_\lambda$, our algorithm based on a $\lambda$-learner. In all experiments except the sparse structure, the minimisation over the alternative sets is equivalent to minimising on the union of a small number of convex sets. For each such convex set, we implement one expert running Follow-The-Leader, and the $\lambda$-learner aggregates these experts using AdaHedge. For the sparse setting, there is still a decomposition into a finite number of convex sets, but that number is combinatorially large and we did not implement SP$_\lambda$ in that case.
\item The OSSB algorithm of~\citep{combes2017minimal}, which solves a plug-in estimate of optimisation problem~\eqref{eq:asymptotic.rate} with empirical estimates and tracks its solution.
\item The vanilla UCB algorithm of~\citep{auer2002finite,garivier2011kl}, which selects the arm with the highest upper confidence bound calculated without additional structural information.
\end{itemize}
The implementation of our algorithms departs from the theory in two ways. First, we set $\varepsilon_t \propto 1/n_t^{j_t}$, where $n_t^{j_t}$ is the number of exploration steps with the current estimated best arm. This is a faster decrease than allowed by our proof. Second, the width of the small confidence intervals~\eqref{eq:small.CI} is smaller: $\ln(n_t)$ instead of $\ln(n_t)+\bigO(\ln\ln t)$.

We also compare with structure specific algorithms: 
\begin{itemize}[nosep]
\item The structural UCB algorithm, denoted by $\mathcal{M}$-UCB, which looks for the highest upper confidence bound over the intersection of the confidence region and the structure. In the linear case, it uses an adapted estimation of the means, such that it is the same as LinUCB~\citep{auer2002using}.
\item The CATSE algorithm of~\citep{jedor2019categorized} for categorised instances, which eliminates all arms in the ``bad'' category once we are confident to tell which category is ``bad''.
\item The ``End of Optimism'' algorithm of~\citep{lattimore2017end} (called "EOO" on figures) for linear instances, which is similar to OSSB with different forced exploration.
\item The OSUB algorithm of~\citep{combes2014unimodal} for unimodal instances, which pulls among the neighbourhood of the empirical optimal arm.
\item The SparseUCB algorithm of~\citep{kwon2017sparse} for sparse instances, which constructs a set of $\geq s$ arms which are estimated to have means larger than $\gamma$ and then play UCB among this set.
\end{itemize}

Figure~\ref{fg:exp} reports the mean regret of these algorithms over 200 repetitions. The shaded areas show an interval around the mean of width twice its empirical standard deviation.  We contrast the empirical regret of the algorithms with the unconstrained lower bound (called ``uncstrd lbd'' in the figures) $\sum_{k\neq i^*(\vmu)}\Delta_k/d(\mu^k, \mu^*)\ln(T)$ and the structural lower bound (called ``lbd'') $V^{\mathcal M}(\vmu)\ln(T)$.

We investigate when the theoretically motivated SP$_k$, SP$_\lambda$ and OSSB also benefit from good empirical performances.
OSSB adapts to the structure in more experiments (the slope of its regret is close to the slope of the structural lower bound, see Figures~\ref{fg:unconstrained1}, \ref{fg:categorized2}, \ref{fg:unimodal1}, \ref{fg:sparse2}), but it incurs a high initial regret in cases~\ref{fg:unconstrained1}, \ref{fg:categorized1}, \ref{fg:categorized2}, \ref{fg:unimodal2}. In Figure~\ref{fg:sparse1}, OSSB performs poorly and is equivalent to Follow-The-Leader, since its exploration test never triggers on that structure. In contrast, our algorithms are safer, in the sense that their regret is at worst comparable to that of UCB, but they adapt to the structure often only later, sometimes after the chosen horizon. We verify that adaptation to the structure indeed happens, notably in Figures~\ref{fg:categorized1} and~\ref{fg:sparse2}.

We are not able to provide a characterisation of the problems on which these algorithms manage to adapt to the structure for small horizons. Obtaining such a characterization, or methods that would adapt to the structure for small times, is the most obvious open problem for structured bandits.

\section{Open Questions in Structured Bandits}

While our algorithms are asymptotically optimal, the dependence of the non-asymptotic bound on problem parameters like the number of arms is definitely not the best possible. Some structures would allow an algorithm to face a problem with a huge number of arms and not suffer from that multiplicity, while our bounds have terms linear in $K$. Broadly, our algorithm is suited to large times, while adaptivity to structure in a small horizon regime remains to be explored.

On the topic of computational complexity, we remarked that our algorithm never solves the lower bound problem completely, but for example in the case of the $k$-learner only computes the best-response over an alternative set. If the structure set $\mathcal M$ is complicated, that computation can still be expensive, or even infeasible. However one could argue that it does not make sense to compute exactly the best response to a noisy problem, and indeed our algorithm remains asymptotically optimal if that minimization is approximate, with an error at the $n^{\text{th}}$ exploration step of order $1/\sqrt{n}$.

A weakness of the analysis of algorithms based on an explore/exploit test like ours is the concentration inequality used to bound the number of times that the exploitation phase is wrongly entered. A concentration result gives a threshold $\beta(t,\delta)$ such that with probability $1- \delta$, a deviation is lower than $\beta(t,\delta)$ for all times. That threshold is linear in the number of arms $K$, while it could be smaller for many particular structures. For example, in the unconstrained case, it can be made to depend only on $\ln(K)$. (see also the discussion on the Pure Exploration Problem Rank by \citet{mixmart}). Adapting the concentration inequality to the structure is in general an open question.

 Finally, the behaviour of our algorithms in experiments has an initial phase where the structure information cannot yet be used. The length of this phase depends on the instance complexity (our bounds currently admit an exponential dependence here), and in some of our experiments we have not decidedly left the initial phase. To shorten this phase without changing the high-level approach, one could work on improving confidence regions for the gradients. Our current choice, $\UCB_t^k$ in \eqref{eq:UCBs}, is possibly quite conservative.

\section*{Acknowledgements}
This work was funded in part by the French government under management of Agence Nationale de la Recherche as part of the ``Investissements d'avenir'' program, reference ANR-19-P3IA-0001 (PRAIRIE 3IA Institute). H. Shao is supported in part by the National Science Foundation under grant CCF-181501. H. Shao would like to thank Emilie Kaufmann for her early discussion on unimodal bandits.

%
%
\bibliography{bib,../bib}

\begin{thebibliography}{27}
\providecommand{\natexlab}[1]{#1}
\providecommand{\url}[1]{\texttt{#1}}
\expandafter\ifx\csname urlstyle\endcsname\relax
  \providecommand{\doi}[1]{doi: #1}\else
  \providecommand{\doi}{doi: \begingroup \urlstyle{rm}\Url}\fi

\bibitem[Abbasi-Yadkori et~al.(2011)Abbasi-Yadkori, P{\'a}l, and
  Szepesv{\'a}ri]{abbasi2011improved}
Abbasi-Yadkori, Y., P{\'a}l, D., and Szepesv{\'a}ri, C.
\newblock Improved algorithms for linear stochastic bandits.
\newblock In \emph{Advances in Neural Information Processing Systems}, pp.\
  2312--2320, 2011.

\bibitem[Agrawal(1995)]{agrawal1995sample}
Agrawal, R.
\newblock Sample mean based index policies by o (log n) regret for the
  multi-armed bandit problem.
\newblock \emph{Advances in Applied Probability}, 27\penalty0 (4):\penalty0
  1054--1078, 1995.

\bibitem[Auer(2002)]{auer2002using}
Auer, P.
\newblock Using confidence bounds for exploitation-exploration trade-offs.
\newblock \emph{Journal of Machine Learning Research}, 3\penalty0
  (Nov):\penalty0 397--422, 2002.

\bibitem[Auer et~al.(2002)Auer, Cesa-Bianchi, and Fischer]{auer2002finite}
Auer, P., Cesa-Bianchi, N., and Fischer, P.
\newblock Finite-time analysis of the multiarmed bandit problem.
\newblock \emph{Machine learning}, 47\penalty0 (2-3):\penalty0 235--256, 2002.

\bibitem[Boyd \& Vandenberghe(2004)Boyd and Vandenberghe]{cvxbook}
Boyd, S. and Vandenberghe, L.
\newblock \emph{Convex Optimization}.
\newblock Cambridge University Press, New York, NY, USA, 2004.
\newblock ISBN 0521833787.

\bibitem[Bubeck \& Cesa-Bianchi(2012)Bubeck and Cesa-Bianchi]{bubeck2012regret}
Bubeck, S. and Cesa-Bianchi, N.
\newblock Regret analysis of stochastic and nonstochastic multi-armed bandit
  problems.
\newblock \emph{Foundations and Trends{\textregistered} in Machine Learning},
  5\penalty0 (1):\penalty0 1--122, 2012.

\bibitem[Cesa-Bianchi \& Lugosi(2012)Cesa-Bianchi and
  Lugosi]{cesa2012combinatorial}
Cesa-Bianchi, N. and Lugosi, G.
\newblock Combinatorial bandits.
\newblock \emph{Journal of Computer and System Sciences}, 78\penalty0
  (5):\penalty0 1404--1422, 2012.

\bibitem[Chen et~al.(2017)Chen, Gupta, Li, Qiao, and Wang]{pmlr-v65-chen17a}
Chen, L., Gupta, A., Li, J., Qiao, M., and Wang, R.
\newblock Nearly optimal sampling algorithms for combinatorial pure
  exploration.
\newblock In \emph{Proceedings of the 2017 Conference on Learning Theory},
  volume~65, pp.\  482--534. PMLR, July 2017.

\bibitem[Combes \& Proutiere(2014)Combes and Proutiere]{combes2014unimodal}
Combes, R. and Proutiere, A.
\newblock Unimodal bandits: Regret lower bounds and optimal algorithms.
\newblock In \emph{International Conference on Machine Learning}, pp.\
  521--529, 2014.

\bibitem[Combes et~al.(2017)Combes, Magureanu, and
  Proutiere]{combes2017minimal}
Combes, R., Magureanu, S., and Proutiere, A.
\newblock Minimal exploration in structured stochastic bandits.
\newblock In \emph{Advances in Neural Information Processing Systems}, pp.\
  1763--1771, 2017.

\bibitem[Dani et~al.(2008)Dani, Hayes, and Kakade]{dani2008stochastic}
Dani, V., Hayes, T.~P., and Kakade, S.~M.
\newblock Stochastic linear optimization under bandit feedback.
\newblock In \emph{21st Annual Conference on Learning Theory - {COLT} 2008,
  Helsinki, Finland, July 9-12, 2008}, pp.\  355--366. Omnipress, 2008.

\bibitem[De~Rooij et~al.(2014)De~Rooij, Van~Erven, Gr{\"u}nwald, and
  Koolen]{de2014follow}
De~Rooij, S., Van~Erven, T., Gr{\"u}nwald, P.~D., and Koolen, W.~M.
\newblock Follow the leader if you can, hedge if you must.
\newblock \emph{The Journal of Machine Learning Research}, 15\penalty0
  (1):\penalty0 1281--1316, 2014.

\bibitem[Degenne \& Koolen(2019)Degenne and Koolen]{multiple.answers}
Degenne, R. and Koolen, W.~M.
\newblock Pure exploration with multiple correct answers.
\newblock In \emph{Advances in Neural Information Processing Systems (NeurIPS)
  32}, pp.\  14564--14573. Curran Associates, Inc., December 2019.

\bibitem[Degenne et~al.(2019)Degenne, Koolen, and M{\'e}nard]{degenne2019non}
Degenne, R., Koolen, W.~M., and M{\'e}nard, P.
\newblock Non-asymptotic pure exploration by solving games.
\newblock In \emph{Advances in Neural Information Processing Systems}, 2019.

\bibitem[Garivier \& Capp{\'e}(2011)Garivier and Capp{\'e}]{garivier2011kl}
Garivier, A. and Capp{\'e}, O.
\newblock The {KL-UCB} algorithm for bounded stochastic bandits and beyond.
\newblock In \emph{Proceedings of the 24th annual conference on learning
  theory}, pp.\  359--376, 2011.

\bibitem[Garivier \& Kaufmann(2016{\natexlab{a}})Garivier and Kaufmann]{GK16}
Garivier, A. and Kaufmann, E.
\newblock Optimal best arm identification with fixed confidence.
\newblock In \emph{Proceedings of the 29th Conference On Learning Theory
  (COLT)}, 2016{\natexlab{a}}.

\bibitem[Garivier \& Kaufmann(2016{\natexlab{b}})Garivier and
  Kaufmann]{garivier2016optimal}
Garivier, A. and Kaufmann, E.
\newblock Optimal best arm identification with fixed confidence.
\newblock In \emph{Conference on Learning Theory}, pp.\  998--1027,
  2016{\natexlab{b}}.

\bibitem[Graves \& Lai(1997)Graves and Lai]{GravesLai97}
Graves, T.~L. and Lai, T.~L.
\newblock Asymptotically efficient adaptive choice of control laws in
  controlled {M}arkov chains.
\newblock \emph{SIAM Journal on Control and Optimization}, 35(3):\penalty0
  715--743, 1997.

\bibitem[Jedor et~al.(2019)Jedor, Perchet, and Louedec]{jedor2019categorized}
Jedor, M., Perchet, V., and Louedec, J.
\newblock Categorized bandits.
\newblock In \emph{Advances in Neural Information Processing Systems}, pp.\
  14399--14409, 2019.

\bibitem[Kaufmann \& Koolen(2018)Kaufmann and Koolen]{mixmart}
Kaufmann, E. and Koolen, W.~M.
\newblock Mixture martingales revisited with applications to sequential tests
  and confidence intervals.
\newblock Preprint, October 2018.

\bibitem[Kveton et~al.(2015)Kveton, Wen, Ashkan, and
  Szepesv{\'{a}}ri]{kveton2015tight}
Kveton, B., Wen, Z., Ashkan, A., and Szepesv{\'{a}}ri, C.
\newblock Tight regret bounds for stochastic combinatorial semi-bandits.
\newblock In \emph{Artificial Intelligence and Statistics}, pp.\  535--543,
  2015.

\bibitem[Kwon et~al.(2017)Kwon, Perchet, and Vernade]{kwon2017sparse}
Kwon, J., Perchet, V., and Vernade, C.
\newblock Sparse stochastic bandits.
\newblock In \emph{Conference On Learning Theory}, 2017.

\bibitem[Lai \& Robbins(1985)Lai and Robbins]{lai1985asymptotically}
Lai, T.~L. and Robbins, H.
\newblock Asymptotically efficient adaptive allocation rules.
\newblock \emph{Advances in applied mathematics}, 6\penalty0 (1):\penalty0
  4--22, 1985.

\bibitem[Lattimore \& Szepesv{\'{a}}ri(2017)Lattimore and
  Szepesv{\'{a}}ri]{lattimore2017end}
Lattimore, T. and Szepesv{\'{a}}ri, C.
\newblock The end of optimism? an asymptotic analysis of finite-armed linear
  bandits.
\newblock In \emph{Artificial Intelligence and Statistics}, pp.\  728--737,
  2017.

\bibitem[Lattimore \& Szepesv{\'a}ri(2019)Lattimore and
  Szepesv{\'a}ri]{lattimore2019bandit}
Lattimore, T. and Szepesv{\'a}ri, C.
\newblock Bandit algorithms.
\newblock \emph{Cambridge University Press}, 2019.

\bibitem[Magureanu et~al.(2014)Magureanu, Combes, and
  Proutiere]{magureanu2014lipschitz}
Magureanu, S., Combes, R., and Proutiere, A.
\newblock Lipschitz bandits: Regret lower bound and optimal algorithms.
\newblock In \emph{Conference on Learning Theory}, pp.\  975--999, 2014.

\bibitem[Thompson(1933)]{thompson1933likelihood}
Thompson, W.~R.
\newblock On the likelihood that one unknown probability exceeds another in
  view of the evidence of two samples.
\newblock \emph{Biometrika}, 25\penalty0 (3/4):\penalty0 285--294, 1933.

\end{thebibliography}
\bibliographystyle{icml2020_style/icml2020}

\appendix
\onecolumn
%
%
%
%

\section{Concentration Results}\label{app:concentration}

We need two concentration results:
\begin{itemize}
	\item One to bound the probability that $\hat{\theta}_t$ has a different best arm than $\theta$ and determine the rule used to stop exploration and switch to exploitation.
	\item The other to build optimistic estimates for the gains that we feed to one of the learners.
\end{itemize}

\subsection{Exploration Stopping Rule}

Our algorithm optimizes the likelihood ratio $\inf_{\lambda\in \neg i_t} \sum_{k=1}^K N_t^k d(\hat{\theta}_t^k, \lambda^k)$. One can then get a concentration inequality of the form: with probability $1- \delta$, $\sum_{k=1}^K N_t^k d(\hat{\theta}_t^k, \theta^k) \le \beta(t,\delta)$ for some function $\beta(t,\delta)$. Then if $\inf_{\lambda\in \neg i_t} \sum_{k=1}^K N_t^k d(\hat{\theta}_t^k, \lambda^k)>\beta(t,\delta)$, we can conclude that $\theta\notin \neg i^*(\hat{\theta}_t)$.

Let $W_{-1}$ be the negative branch of the Lambert W function and for $x>1$, let $\overline{W}(x) = -W_{-1}(-e^{-x})$. It verifies $x + \log(x) \le \overline{W}(x) \le x + \log(x) + \min\{\frac{1}{2}, \frac{1}{\sqrt{x}}\}$.
\begin{theorem}\label{thm:explore_exploit_threshold}
Let $t>1$, $\theta\in \mathcal M$. With probability $1- \delta$,
\begin{align*}
\sum_{k=1}^K N_t^k d(\hat{\theta}_t^k, \theta^k) \leq \beta(t,\delta)
\quad \text{with} \quad
\beta(t,\delta) = 2K \overline{W}\left( \frac{1}{2K}\log\frac{e}{\delta} + \frac{1}{2}\log(8eK\log t) \right) \: .
\end{align*}
\end{theorem}

\begin{proof}
From the proof of Theorem 2 in \cite{magureanu2014lipschitz}, if we restrict ourselves to $t>1$, we obtain for any $x>K+1$,
\begin{align*}
\mathbb{P}(\sum_{k=1}^K N_t^k d(\hat{\theta}_t^k, \theta^k) > x)
\le e\left( \frac{2 e x^2 \log(t)}{K} \right)^K e^{-x} \: .
\end{align*}
We can equivalently write that, for any $t>1$, with probability $1- \delta$,
\begin{align*}
\sum_{k=1}^K N_t^k d(\hat{\theta}_t^k, \theta^k)
\le \beta(t,\delta) \: ,
\end{align*}
where $\beta(t,\delta)$ is the solution in $x$ of $\delta = e\left( \frac{2 e x^2 \log(t)}{K} \right)^K e^{-x}$ .
\end{proof}

\subsection{Confidence Intervals}

\begin{theorem}\label{thm:maximal_concentration_inequality}
For all $t>1$ and $\eta>0$ such that $\log_{1+\eta}(t)$ is not an integer, with probability $1-\delta$, for all $s\le t$ and all $k\in [K]$,
\begin{align*}
N_s^k d(\hat{\theta}_s^k, \theta^k)
&\le (1+\eta)\left( \log\frac{2K\log t}{\delta}-\log\log(1+\eta) \right) \: .
\end{align*}
Optimizing over $\eta$, we obtain that with probability $1-\delta$, for all $s\le t$ and all $k\in [K]$,
\begin{align*}
N_s^k d(\hat{\theta}_s^k, \theta^k)
&\le \overline{W}(\log\frac{2K\log t}{\delta})\exp\left(1/\overline{W}(\log\frac{2K\log t}{\delta})\right) \: .
\end{align*}
For all $\eta>0$, with probability $1- \delta$, for all $s\in \mathbb{N}$ and all $k\in[K]$, if $N_s^k>0$ then
\begin{align*}
N_s^k d(\hat{\theta}_s^k, \theta^k)
&\le (1+\eta)\left( \log\frac{2\zeta(2)K}{\delta} - 2\log(1 + \frac{\log N_s^k}{\log(1+\eta)}) \right) \: .
\end{align*}
\end{theorem}

\begin{proof}
We prove the inequality for one arm, for one-sided deviations. The final result is then obtained by an union bound over arms and the two sides. We drop the arm superscript in this proof.

For $i\in\{0,\ldots,\lceil \log_{1+\eta}(t) \rceil\}$, let $\theta_i$ be such that $d(\theta_i, \theta)= \frac{x}{(1+\eta)^i}$. We denote the quantity $\lceil \log_{1+\eta}(t) \rceil$ by $m_t$.

Let $i_t\in\{0, \ldots, m_t-1\}$ be such that $(1+\eta)^i \le N_t < (1+\eta)^{i+1}$. Remark that 
\begin{align*}
\{\hat{\theta}_t \ge \theta, \: N_t d(\hat{\theta}_t, \theta)\ge x\}
&\subseteq \{\hat{\theta}_t \ge \theta, \: d(\hat{\theta}_t, \theta)\ge \frac{x}{(1+\eta)^{i_t+1}}\}
\\
&= \{\hat{\theta}_t \ge \theta, \: d(\hat{\theta}_t, \theta) - d(\hat{\theta}_t, \theta_{i_t+1}) \ge \frac{x}{(1+\eta)^{i_t+1}}\}
\\
&\subseteq \{\hat{\theta}_t \ge \theta, \: N_t(d(\hat{\theta}_t, \theta) - d(\hat{\theta}_t, \theta_{i_t+1})) \ge \frac{x}{(1+\eta)}\} \: .
\end{align*}
The equality uses that for $\lambda>\theta$, for one-dimensional exponential families, $(\hat{\theta}_t \ge \theta, \: d(\hat{\theta}_t, \theta)\ge d(\lambda, \theta)) \Leftrightarrow (\hat{\theta}_t \ge \theta, \: d(\hat{\theta}_t, \theta) - d(\hat{\theta}_t, \lambda)\ge d(\lambda, \theta))$ .

For all distributions $\rho$ on $\mathbb{R}$, we can verify that the following quantity is a martingale with expectation 1:
\begin{align*}
W_t(\rho)
= \mathbb{E}_{y\sim \rho}e^{N_t (d(\hat{\theta}_t, \theta) - d(\hat{\theta}_t, y))}
\: .
\end{align*}
Let $\rho$ be the uniform distribution on $\{\theta_1,\ldots, \theta_{m_t}\}$. Summing up the preceding arguments and using Doob's inequality,
\begin{align*}
\mathbb{P}(\hat{\theta}_t \ge \theta, \: N_t d(\hat{\theta}_t, \theta)\ge x )
&\le \mathbb{P}(\hat{\theta}_t \ge \theta, \: N_t(d(\hat{\theta}_t, \theta) - d(\hat{\theta}_t, \theta_{i_t+1})) \ge \frac{x}{(1+\eta)})
\\
&\le \mathbb{P}(\hat{\theta}_t \ge \theta, \: \log \sum_i \frac{1}{m_t-1}e^{N_t(d(\hat{\theta}_t, \theta) - d(\hat{\theta}_t, \theta_{i}))} \ge \frac{x}{(1+\eta)} - \log (m_t-1))
\\
&= \mathbb{P}(\hat{\theta}_t \ge \theta, \: W_t(\rho) \ge \frac{1}{m_t-1}e^{\frac{x}{(1+\eta)}})
\\
&\le \frac{\log t}{\log(1+\eta)} e^{-x/(1+\eta)}
\end{align*}
Equivalently, with probability $1- \delta$, if $\hat{\theta}_t \ge \theta$ then
\begin{align*}
N_t d(\hat{\theta}_t, \theta)
\le (1+\eta)\left(\log\frac{\log t}{\delta} - \log\log(1+\eta)\right) \: .
\end{align*}

We now prove the third claim.

Extend the definition of $\theta_i$ to all $i\in \mathbb{N}$. With $\rho$ the distribution supported on $\cup_{i\ge 1}\{\theta_i\}$ with $\rho(\theta_i) = 1/(\zeta(2)i^2)$, we get
\begin{align*}
\{\hat{\theta}_t \ge \theta, \: N_t d(\hat{\theta}_t, \theta)\ge x \}
&\subseteq \{\hat{\theta}_t \ge \theta, \: N_t(d(\hat{\theta}_t, \theta) - d(\hat{\theta}_t, \theta_{i_t+1})) \ge \frac{x}{(1+\eta)}  \}
\\
&\subseteq \{\hat{\theta}_t \ge \theta, \: \log \sum_i \rho(\theta_{i})e^{N_t(d(\hat{\theta}_t, \theta) - d(\hat{\theta}_t, \theta_{i}))} \ge \frac{x}{(1+\eta)} + \log \rho(\theta_{i_t+1}) \}
\\
&= \{\hat{\theta}_t \ge \theta, \: W_t(\rho) \ge \rho(\theta_{i_t+1})e^{\frac{x}{(1+\eta)}} \}
\: .
\end{align*}
With probability $1- \delta$, the martingale $W_t(\rho)$ is smaller than $1/\delta$. That is, with probability $1- \delta$,
\begin{align*}
N_t d(\hat{\theta}_t, \theta)
&\le (1+\eta)\left( \log\frac{1}{\delta} - \log \rho(\theta_{i_t+1}) \right)
\\
&=   (1+\eta)\left( \log\frac{1}{\delta} + \log (\zeta(2)(i_t+1)^2) \right)
\\
&\le (1+\eta)\left( \log\frac{1}{\delta} + \log (\zeta(2)(\log_{1+\eta}(N_t)+1)^2) \right)
\: .
\end{align*}
\end{proof}


\section{Tracking}\label{app:tracking}

We call tracking the following interaction. Starting from vectors $W_0 = N_0 = (0, \ldots, 0) \in \mathbb{R}^K$, for each stage $t=1,2,\ldots$
\begin{itemize}
	\item Nature reveals a vector $w_t$ in the simplex $\triangle_K$ and updates $W_t = W_{t-1} + w_t$.
	\item A tracking rule selects $k_t \in [K]$ based on the sequence $(w_1, \ldots, w_t)$ and forms $N_t = N_{t-1} + e_{k_t}$, where $(e_i)_{i\in [K]}$ are the canonical basis.
\end{itemize}
Note that $w_t$ is known to the tracking rule when choosing $k_t$. We are interested in tracking rules such that $\Vert N_t - W_t\Vert_{\infty}$ is as small as possible.

\begin{definition}
We call C-Tracking any rule which for all stages $t\ge 1$ selects $k_t \in \argmin_{k\in [K]} N_{t-1}^k - W_t^k$.
\end{definition}
This defines C-tracking up to the choice of $k_t$ when the argmin is not unique.

\begin{theorem}\label{thm:tracking}
The C-Tracking procedure described above ensures that for all $t\in \N$, for all $k\in [K]$,
\begin{align*}
-\sum_{j=2}^K \frac{1}{j} \le N_t^k - W_t^k \le 1 \: .
\end{align*}
\end{theorem}
\begin{proof}
The upper bound was proved in \cite{GK16}. We prove the lower bound.

Let $\mathcal S_0 = \{v \in \mathbb{R}^K \: : \: \sum_{k=1}^K v^k = 0\}$. The tracking procedure is such that for all stages $t\in\N$, $N_t-W_t \in S_0$. Our proof strategy is to characterize the subset of $\mathcal S_0$ that can be reached during the tracking procedure, starting from $\v_0 = N_0-W_0 = \zeros$ .

We define a move $\rightarrow_\w$ as function from $\mathcal{S}_0$ to itself parametrized by $\w$ that maps $\v$ to $\v - \w + \e_k$, where $k=\argmin_{j\in[K]} \v - \w$. If the value of that function at $\v$ is $\u$, we write $\v \rightarrow_\w \u$ .
A vector $\u \in \mathcal{S}_0$ is said to be reachable in one move from $\v \in \mathcal{S}_0$ if there exists $\w \in \triangle_K$ such that $\v \rightarrow_\w \u$. We denote it by $\v \rightarrow \u$. It is said to be reachable from $\v$ if there is a finite sequence of such moves such that $\v \rightarrow \ldots \rightarrow \u$.

A reverse move $\leftarrow_{k,\w}$ is a function from $\mathcal{S}_0$ to itself parametrized by $k$ and $\w$ that maps $\v$ to $\v + \w - \e_k$. A reverse move is said to be valid at $\v$ if $v^k \leq \min_j v^j + 1$. If the value of that function at $\v$ is $\u$, we write $\u \leftarrow_{k,\w} \v$ .
A vector $\u \in \mathcal{S}_0$ is said to be reverse-reachable in one move from $\v \in \mathcal{S}_0$ if there exists $k\in[K]$ and $\w \in \triangle_K$ such that $\u \leftarrow_{k,\w} \v$ and such that this is a valid reverse move at $\v$. We denote it by $\u \leftarrow \v$.

We now prove that $(\u \rightarrow \v) \Leftrightarrow (\u \leftarrow \v)$ . First, if $\u \rightarrow \v$ then let $\w$ be the parameter of a move $\u \rightarrow_\w \v$ and let $k = \argmin_{j\in[K]} \u - \w$. Then $\u \leftarrow_{k,\w} \v$ is a valid reverse move. Second, if $\u \leftarrow_{k,\w} \v$ is a valid reverse move, then $k = \argmin_{j\in[K]} \u - \w$ and we have $\u \rightarrow_\w \v$ .

We characterize the elements $\v$ of $\mathcal S_0$ such that $\zeros \leftarrow \ldots \leftarrow \v$ .

Let $\u,\v \in \mathcal{S}_0$ be such that $\u \leftarrow \v$. Let $M_\v = \{k\in[K]: v^k \le \min_j v^j + 1\}$. Then for any set $S\subseteq[K]$ such that $M_\v\subseteq S$, $\sum_{i\in S} u^i \le \sum_{i\in S} v^i$. Indeed, for the reverse move to be valid, one of the coordinates in $M_\v$ was decreased by 1, and they were added coordinates of a $\w\in \triangle_K$, that sum at most to 1.

Let $S\subseteq [K]$ and $A_S = \{u\in\mathcal{S}_0 : \forall k \notin S, \: u^k > \frac{1}{|S|}\sum_{i\in S}u^i + 1\}$. We now prove that if $\u \leftarrow \v$ and $\v \in A_S$, then $\u \in A_S$ and as a consequence, that if $\u \leftarrow \ldots \leftarrow \v$ and $\v \in A_S$, then $\u \in A_S$. Indeed,
\begin{itemize}[nosep]
	\item Since $\v \in A_S$, we have $M_\v \subseteq S$, hence the previous remark proves $\frac{1}{|S|}\sum_{i\in S}u^i \leq \frac{1}{|S|}\sum_{i\in S}v^i$.
	\item For $k\notin S$, then $k\notin M_\v$ and $u^k \geq v^k > \frac{1}{|S|}\sum_{i\in S}v^i + 1 \geq \frac{1}{|S|}\sum_{i\in S}u^i + 1$ .
\end{itemize}

Since $\zeros \notin \bigcup_{S \in \mathcal{P}([K])\setminus\{[K]\}} A_S$, we can now state that if $\zeros \leftarrow \ldots \leftarrow \v$, then $\v \notin \bigcup_{S \in \mathcal{P}([K])\setminus\{[K]\}} A_S$ .

Let $j\in[2:K]$ and let $\v_{(j)} \in \mathcal{S}_0$ be such that $v_{(j)}^1 \ge \ldots \ge v_{(j)}^{j-1} > v_{(j)}^j = \ldots = v_{(j)}^K$. Then we will prove that one of the two following statements is true:
\begin{enumerate}
\item $v_{(j)}^{j-1} > v_{(j)}^j + 1$ and $\v_{(j)}$ is not reachable from $\zeros$,
\item $v_{(j)}^{j-1} \le v_{(j)}^j + 1$. Then let $v_{j-1:K}$ be the mean of $v_{(j)}^{j-1},\ldots,v_{(j)}^K$ and let $\v_{(j-1)}$ be the vector with $v_{(j-1)}^1 = v_{(j)}^1$, ..., $v_{(j-1)}^{j-2} = v_{(j)}^{j-2}$ and $v_{(j-1)}^{j-1} = \ldots = v_{(j-1)}^K = v_{j-1:K}$. Then $\v_{(j-1)} \leftarrow \v_{(j)}$ .
\end{enumerate}
Case 1: $v_{(j)}^{j-1} > v_{(j)}^j + 1$. Let $S = [j:K]$. then $S\neq[K]$ and $\v\in A_S$. Hence $\v$ is not reachable from $\zeros$.

Case 2: $v_{(j)}^{j-1} \le v_{(j)}^j + 1$. Let $\w$ be defined by $w^1 = \ldots = w^{j-2} = 0$, $w^{j-1} = 1 - (K-j+1)(v_{j-1:K} - v_{(j)}^j)$ and $w^j = \ldots = w^K = v_{j-1:K} - v_{(j)}^j$. Since $v_{(j)}^{j-1} \le v_{(j)}^j + 1$ and $\w\in\triangle_K$, the reverse move $\leftarrow_{j-1,\w}$ is valid at $\v_{(j)}$. Then the image of $\v_{(j)}$ by that reverse move is $\v_{(j-1)}$. Note that $w^{j-1} \ge w^j = \ldots = w^K$.

We have now all the tools to state the characterization of the the elements $\v$ of $\mathcal S_0$ such that $\zeros \leftarrow \ldots \leftarrow \v$ . By a simple induction using the last case distinction, we have the following:
let $\v \in \mathcal S_0$ and let $i_1, \ldots, i_K\in[K]$ be such that $v^{i_1} \ge \ldots \ge v^{i_K}$. If $\zeros \leftarrow \ldots \leftarrow \v$, then there exists $\u_1, \ldots, \u_{K-2}$ and $\w_1, \ldots, \w_{K-1}$ such that
\begin{enumerate}
\item $\zeros \leftarrow _{i_1,\w_1} \u_1 \leftarrow_{i_2,\w_2} \ldots \leftarrow_{i_{K-2},\w_{K-2}} \u_{K-2} \leftarrow_{i_{K-1}, \w_{K-1}} \v$,
\item for all $j \in [K-2]$, $u_j^{i_1} \ge \ldots u_j^{i_j} \ge u_j^{i_{j+1}} = \ldots = u_j^{i_K} = \frac{1}{K-j}\sum_{k=j+1}^K v^{i_k}$ ,
\item for all $j \in [K-1]$, $w_j^{i_1} = \ldots w_j^{i_{j-1}} = 0$ and $w_j^j \ge w_j^{i_{j+1}} = \ldots = w_j^{i_K}$. 
\end{enumerate}
In order to prove the theorem, we then only need a bound on $v^{i_K} = - \sum_{j=1}^{K-1} w_j^K$. The characterization of $\w_j$ implies that $w_j^K \leq 1/(K-j+1)$ . Hence $v^{i_K} \ge - \sum_{j=2}^{K} \frac{1}{j}$ .

\end{proof}

\begin{lemma}
For all tracking rules, for all stages $t$ and vectors $N_t$, $W_t$, there exists a sequence $w_{t+1}, \ldots, w_{t+K-1} \in \triangle_{K-1}$ such that $\min_{k\in[K]} (N_{t+K-1}^k - W_{t+K-1}^k) - (N_{t}^k - W_{t}^k) \le -\sum_{j=2}^K \frac{1}{j}$ .

In particular, with $t=0$ and $N_0=W_0=0$, there exists a sequence $w_{1}, \ldots, w_{K-1} \in \triangle_{K-1}$ such that $\min_{k\in[K]} (N_{K-1}^k - W_{K-1}^k) = -\sum_{j=2}^K \frac{1}{j}$ .
\end{lemma}
\begin{proof}
Let $w_{t+1}$ be the vector with all coordinates equal to $1/K$.
Define then $w_{t+2}^k = 1/(K-1)$ for $k\neq k_{t+1}$ and $w_{t+2}^{k_{t+1}} = 0$.

In general, let $n_s$ be the number of different arms chosen between times $t+1$ and $t+s$ and let $w_{t+s}^k = 1/(K-n_s)$ if $k$ was not yet chosen in that time interval, and $w_{t+s}^k = 0$ otherwise.

There is one arm $i\in[K]$ such that $i\neq k_{t+1}, \ldots, i\neq k_{t+K-1}$. For that arm, $w_{t+1}^i+\ldots+w_{t+K-1}^i \ge \sum_{j=2}^K \frac{1}{j}$ (with equality if all $k_{t+s}$ are different) and $N_{t+K-1}^i = N_t^i$.
\end{proof}


\section{Regret Bound Proof}\label{app:regret_proof}

\subsection{Assumptions}

We summarize below the assumptions under which our results are proven. We show in Appendix~\ref{app:verify_assumption} that all structures used in the paper verify  Assumption~\ref{ass:neg_is_lip}.

\begin{assumption}
Each arm has a distribution in the same 1-parameter exponential family, with parameter in an open interval $\Theta\subseteq \mathbb{R}$. The distribution of arm $k$ has mean parameter $\mu^k$.
\end{assumption}

\begin{assumption}\label{ass:sub_gaussian}
The arm distributions are $\sigma^2$-sub-Gaussian, meaning that they verify $\log \mathbb{E} e^{t (X - \mathbb{E}[X])} \le \frac{1}{2}\sigma^2 t^2$ for all $t \in \mathbb{R}$.
\end{assumption}

Consequence: the Kullback-Leibler divergence between distributions with means $x$ and $y$ verifies $d(x,y) \ge (x - y)^2/(2 \sigma^2)$~.

\begin{definition}
A structure set is a subset $\mathcal M$ of $\Theta^K$.
\end{definition}

For simplicity we consider only structures with a unique best arm.
\begin{assumption}\label{ass:i*_unique}
For all $\vmu \in \mathcal M$, $i^*(\vmu) = \argmax_k \mu^k$ is uniquely defined.
\end{assumption}

\begin{definition}
Let $\text{cl}(A)$ denote the closure of a set $A$. Given a structure set $\mathcal M \subseteq \Theta^K$, for $i \in [K]$ we define the alternative sets
\begin{align*}
\neg i &= \text{cl}(\{\vlambda\in \mathcal M \: : \: i^*(\vlambda)\neq i\}) \: ,
\\
\neg_x i &= \neg i \cap \{\vxi\in \Theta^K\: : \: \xi^i = x\} \: .
\end{align*}
\end{definition}
Our algorithm minimizes over the sets $\neg i$, while the asymptotic complexity is expressed in terms of the sets $\neg_x i$.

\begin{definition}
For all $\vmu,\vlambda \in  \Theta^K$, we define
\begin{align*}
Q_\vmu(\vlambda) &= \{\vxi \in \Theta^K \: : \: \forall k \in [K], \: d(\mu^k, \xi^k) \ge d(\mu^k, \lambda^k)\} \: ,
\\
Q_\vmu^>(\vlambda) &= \{\vxi \in \Theta^K \: : \: \forall k \in [K], \: d(\mu^k, \xi^k) \ge d(\mu^k, \lambda^k), \exists j \in [K], d(\mu^j, \xi^j) > d(\mu^j, \lambda^j)\} \: ,
\end{align*}
For all $\vmu \in \Theta^K$ and $\Lambda \subset \Theta^K$, we define the boundary of $\Lambda$ relative to $\vmu$ by
\begin{align*}
\partial_\vmu \Lambda = \{\vlambda \in \Lambda \: : \: \forall \vxi \in \Lambda, \vlambda \notin Q_\vmu^>(\vxi)\} \: .
\end{align*}
\end{definition}

The next lemma states that the lower bound problem can be restricted to that boundary (and justifies the term ``boundary'').
\begin{lemma}
For all $\vmu \in \mathcal M$, $\v \in (\mathbb{R}^+)^K$, there exists $\vlambda \in \partial_\vmu \neg i^*(\vmu)$ such that $\vlambda \in \argmin_{\vxi \in \neg i^*(\vmu)} \sum_{k=1}^K v^k d(\mu^k, \xi^k)$.
\end{lemma}
The main idea that we will use is that the lower bound problem can be restricted to $\neg i^*(\vmu)$ minus any union of $Q_\vmu(\vlambda)\setminus\{\vlambda\}$ for some $\vlambda \in \neg i^*(\vmu)$.
We introduce the boundary because the Sparse structure (see Section~\ref{app:verify_assumption}) is such that $\neg i^*(\vmu)$ does not verify the property described in Assumption~\ref{ass:neg_is_lip} (see below), but $\partial_\vmu \neg i^*(\vmu)$ does.

\begin{assumption}\label{ass:M_bounded}
There exists a compact set $\mathcal C \subseteq \Theta^K$ such that $\mathcal M \subseteq \mathcal C$.
\end{assumption}

As a consequence of that assumption, we have the following lemma.
\begin{lemma}\label{lem:bounds_on_d}
There exists $C>0$ such that for all $\theta, \lambda \in \mathcal M$, for all $k\in [K]$, $d(\theta^k,\lambda^k)\le C/2$.
There exists $L>0$ such that for all $\lambda\in \mathcal M$, for all $k\in [K]$, the function $x\mapsto d(x,\lambda^k)$ is $L$-Lipschitz on $\mathcal M$.
There exists $L'>0$ such that for all $\lambda\in \mathcal M$, for all $k\in [K]$, the function $x\mapsto d(\lambda^k,x)$ is $L'$-Lipschitz on $\mathcal M$.
\end{lemma}

The next assumption, which describes a sort of Lipschitz property for the structure, is key to linking the value of the perturbed lower bound problem and the unperturbed one together.

\begin{assumption}\label{ass:neg_is_lip}
There exists $c_{\mathcal M}$ such that for all $\vmu \in \mathcal M$, denoting $i^*(\vmu) = *$, if $\neg_{\mu^*} * \neq \emptyset$ then
\begin{align*}
\forall \vlambda \in \partial_\vmu \neg *, \: \exists \vxi \in \neg_{\mu^*} *, \: \forall k\in[K], \: \vert \xi^k - \lambda^k \vert \le c_{\mathcal M} \vert \mu^* - \lambda^* \vert \: .
\end{align*}
\end{assumption}

\subsection{Events and Counts}
We first define random variables and events used in the analysis.

Events and random variables based on the algorithm definition:
\begin{itemize}[nosep]
\item $E_t$: the algorithm explores at stage $t$. 
\item $E_t^j = E_t \cap \{j_t = j\}$.
\item $E_t^{j\%2} = E_t^j \cap \{n_{t-1}^j \text{ mod } 2 = 1\}$.
\item $n_t = \sum_{s\leq t}\mathbb{I}(E_s)$.
\item $n_t^j = \sum_{s\leq t}\mathbb{I}(E_s^j)$.
\end{itemize}

Events and random variables based on concentration:
\begin{itemize}[nosep]
\item $\mathcal{E}_t = \{\sum_{k=1}^K N_t^k d(\hat{\theta}_t^k, \theta^k) \leq f(t)\}$ where $f(t) = \beta(t, 1/(t\log(t)))$.
\item $c_t = \sum_{s\leq t} \mathbb{I}(\overline{\mathcal{E}_s})$, number of stages without concentration at that level.
\item $\mathcal{E}_n^{exp} = \{\forall k\in[K], \forall t\in\N^*, \: N_{t-1}^k d(\hat{\theta}_{t-1}^k,\theta^k) \leq g_{t-1}(n) \}$ with $g_t(n) = (1+ \eta)[\log(n) + \log(\frac{2K^2\log^2(t)}{\log(1+\eta)})]$ for a fixed $\eta>0$.
\item $m_t = \sum_{s\leq t} \mathbb{I}(E_s \cap \overline{\mathcal{E}_{n_{s-1}^{j_s}}^{exp}})$.
\item $m_t^j = \sum_{s\leq t} \mathbb{I}(E_s^j \cap \overline{\mathcal{E}_{n_{s-1}^j}^{exp}})$.
\end{itemize}

For $k\in[K]$ and $t \in \mathbb{N}$, we define
\begin{align*}
\overline{\theta_{t-1}^k}
&= \sup\{\xi\in \mathbb{R}\: : \: N_{t-1}^k d(\hat{\theta}_{t-1}^k,\xi) \leq g_{t-1}(n_{t-1}^{j_t})\} \: ,
\\
\underline{\theta_{t-1}^k}
&= \inf\{\xi\in \mathbb{R}\: : \: N_{t-1}^k d(\hat{\theta}_{t-1}^k,\xi) \leq g_{t-1}(n_{t-1}^{j_t})\} \: ,
\end{align*}
such that $\mathcal{E}_{n_{t-1}^{j_t}}^{exp} = \{\forall k\in[K], \forall t\in\N^*, \: \hat{\theta}_{t-1} \in [\underline{\theta_{t-1}^k}, \overline{\theta_{t-1}^k}] \}$ .
Due to the sub-Gaussian assumption, we have
\begin{align*}
\hat{\theta}_{t-1}^k - \sqrt{2\sigma^2 \frac{g_{t-1}(n_{t-1}^{j_t})}{N_{t-1}^k}}
\le
\underline{\theta_{t-1}^k}
\le 
\hat{\theta}_{t-1}^k
\le
\overline{\theta_{t-1}^k}
\le
\hat{\theta}_{t-1}^k + \sqrt{2\sigma^2 \frac{g_{t-1}(n_{t-1}^{j_t})}{N_{t-1}^k}}
 \: .
\end{align*}

\paragraph{Expectations of some of these random variables.}
\begin{align*}
\mathbb{E}[c_t]
&\le \sum_{s=1}^t \mathbb{P}(\overline{\mathcal E}_s)
\le 1+\sum_{s=2}^t \frac{1}{s\log s}
\le 1+\log\log(t)\: .
\\
\mathbb{E}[m_t^j]
&\le \sum_{n=1}^{t} \mathbb{P}(\overline{\mathcal{E}_{n}^{exp}})
\le 1+\sum_{n=2}^t \frac{1}{K n\log n}
\le 1+\frac{1}{K}\log\log(t)
\: .
\\
\mathbb{E}[m_t]
&= \sum_{j=1}^K \mathbb{E}[m_t^j]
\le K+\log\log(t)
\: .
\end{align*}

\subsection{Estimation of the Gaps}\label{sec:gap_estimates}

Once a candidate best arm $j_s$ is selected at stage $s$, we form estimates of the gaps for each arm. We use an optimistic estimator for the value of the best arm and define for all $k\in[K]$,
$
\tilde{\Delta}_s^k = \max\left\{\varepsilon_s, \overline{\theta_{s-1}^{j_s}} - \overline{\theta_{s-1}^k} \right\} \: .
$
This quantity estimates the gap in a worst case fashion, with an added perturbation of $\varepsilon_s$.
Note that $\tilde{\Delta}_s^{j_s} = \varepsilon_s$. Obviously for all $k\in[K]$, $\tilde{\Delta}_s^k \geq \varepsilon_s$. Under $\mathcal{E}^{exp}_{n_{s-1}^{j_s}}$, for $k\neq j_s$,
\begin{align}
\tilde{\Delta}_s^k
\ge \max\left\{ \varepsilon_s, \theta^{j_s} - \theta^k - (\overline{\theta_{s-1}^k} - \underline{\theta_{s-1}^k}) \right\}
&\ge \max\left\{ \varepsilon_s, \theta^{j_s} - \theta^k - 2\sqrt{2\sigma^2\frac{g_{s-1}(n_{s-1}^{j_s})}{N_{s-1}^k}} \right\} \: . \label{eq:gap_estimation_lower_bound}
\\
\tilde{\Delta}_s^k
\le \max\left\{ \varepsilon_s, \theta^{j_s} - \theta^k + (\overline{\theta_{s-1}^{j_s}} - \underline{\theta_{s-1}^{j_s}}) \right\}
&\le \max\left\{ \varepsilon_s, \theta^{j_s} - \theta^k + 2\sqrt{2\sigma^2\frac{g_{s-1}(n_{s-1}^{j_s})}{N_{s-1}^{j_s}}} \right\} \: . \label{eq:gap_estimation_upper_bound}
\end{align}

\subsection{UCBs}\label{sec:ucb}

We introduce upper confidence bounds on the ratio of Kullback-Leibler divergence and gap.
\begin{align*}
\UCB_s^k
&= \sup_{\xi \in [\underline{\theta_{s-1}^k}, \overline{\theta_{s-1}^k}]} \frac{\ex_{\vlambda\sim\q_s} d(\xi,  \lambda^{k})}{\max\left\{\varepsilon_s, \indicator\{k\neq j_s\}(\overline{\theta_{s-1}^{j_s}} - \xi)\right\}}
\: .
\end{align*}

Note that the supremum in $\theta$ can only be achieved at either end of the range, or at the kink induced by the $\max$ in the denominator, as a non-negative convex (since divergence in first argument) over positive concave (since linear/constant) function is quasi-convex \citep[Example~3.38]{cvxbook}.

\paragraph{Upper bound.}
\begin{align}
\UCB_s^k
\leq  \frac{\sup_{\xi \in [\underline{\theta_{s-1}^k}, \overline{\theta_{s-1}^k}]} \ex_{\vlambda\sim\q_s} d(\xi,  \lambda^{k})}
{\inf_{\xi \in [\underline{\theta_{s-1}^k}, \overline{\theta_{s-1}^k}]}\max\left\{\varepsilon_s, \indicator\{k\neq j_s\}\left[\overline{\theta_{s-1}^{j_s}} - \xi\right]\right\}}
= \frac{\sup_{\xi \in [\underline{\theta_{s-1}^k}, \overline{\theta_{s-1}^k}]}\ex_{\vlambda\sim\q_s} d(\xi,  \lambda^{k})}{\tilde{\Delta}_s^k} \: . \label{eq:ucb_upper_bound}
\end{align}

\paragraph{Lower bound.} We consider only the case $j_s=*$.

We now relate $\UCB_s^k$ and $\frac{\ex_{\vlambda\sim \q_s} d(\theta^k, \lambda^k)}{\Delta_s^k}$ where $\Delta_s^k = \max\{\varepsilon_s, \Delta^k\}$. Under $\mathcal{E}^{exp}_{n_{s-1}^*}$, $\theta^k$ belongs to the interval over which the maximization is performed in the definition of $\UCB_s^k$ for all $k\in[K]$, hence
\begin{align*}
\UCB_s^k
&\geq \frac{\ex_{\vlambda\sim \q_s} d(\theta^k, \lambda^k)}{\max\{\varepsilon_s, \indicator\{k\neq *\}[\overline{\theta_{s-1}^{j_s}} - \theta^k]\}}
\geq \frac{\ex_{\vlambda\sim \q_s} d(\theta^k, \lambda^k)}{\max\left\{\varepsilon_s, \indicator\{k\neq *\}\left[\Delta^k + 2\sqrt{2\sigma^2\frac{g_{s-1}( n_{s-1}^*)}{N_{s-1}^*}}\right]\right\}}.
\end{align*}
Hence $\UCB_s^* \geq \frac{\ex_{\vlambda\sim \q_s} d(\theta^*, \lambda^*)}{\varepsilon_s}$ and for $k\neq *$, using $1/(1+x) \geq 1-x$,
\begin{align}
\UCB_s^k
&\geq \frac{\ex_{\vlambda\sim \q_s} d(\theta^k, \lambda^k)}{\Delta_s^k} \frac{\Delta_s^k}{\Delta_s^k + 2\sqrt{2\sigma^2\frac{g_{s-1}( n_{s-1}^*)}{N_{s-1}^*}}}
\geq \frac{\ex_{\vlambda\sim \q_s} d(\theta^k, \lambda^k)}{\Delta_s^k}\left(1 - \sqrt{\frac{8\sigma^2g_{s-1}( n_{s-1}^*)}{(\Delta_s^k)^2 N_{s-1}^*}}\right)
\: .\label{eq:ucb_lower_bound}
\end{align}

\subsection{Decomposition of the regret}
The regret at the end of stage $t$ is
\begin{align*}
R_t
= \sum_{s\leq t} \Delta_{k_s}
&= \underbrace{\sum_{s\leq t, \overline{E_s} \cap \overline{\mathcal{E}_s}} \Delta_{k_s}}_{R_t^{(1)}}
+ \underbrace{\sum_{s\leq t, \overline{E_s} \cap \mathcal{E}_s} \Delta_{k_s}}_{R_t^{(2)}}
+ \underbrace{\sum_{s\leq t, E_s \cap \overline{\mathcal{E}_{n_{s-1}^{j_s}}^{exp}}} \Delta_{k_s}}_{R_t^{(3)}}
+ \sum_{j=1}^K \underbrace{\sum_{s\leq t, E_s^j \cap \mathcal{E}_{n_{s-1}^j}^{exp}} \Delta_{k_s}}_{R_t^{(4j)}}\: .
\end{align*}

\subsection{Bounding the three first terms of the regret}\label{sub:easy_regret_terms}
First, by definition of $c_t$,
$\ex R_t^{(1)} \leq \Delta_{\max} \ex c_t$ .

$R_t^{(2)}$ corresponds to the event that concentration holds and we enter the exploitation phase. That is, we have
\begin{align*}
\min_{\vlambda\in\neg k_t} \sum_k N_{t-1}^k d(\hat{\theta}^k_{t-1}, \lambda^k)
~\stackrel{\text{\tiny since $\overline{E_t}$}}{>}~
f(t-1)
~\stackrel{\text{\tiny since $\mathcal{E}_t$}}{\ge}~
\sum_k N_{t-1}^k d(\hat{\theta}^k_{t-1}, \theta^k)
\: .
\end{align*}
In that case, $k_t = *$ (or we would find a contradiction since then $\vtheta \in \neg k_t$) and there is no regret. $R_t^{(2)} = 0$.

Finally, by definition of $m_t$,
$\ex R_t^{(3)}
= \ex \sum_{s\leq t, E_s \cap \overline{\mathcal{E}_{n_{s-1}^{j_s}}^{exp}}} \Delta_{k_s}
\leq \Delta_{\max} \ex m_t$ .

We proved at that point that
\begin{align}
\mathbb{E} R_t
&\le \Delta_{\max}(\mathbb{E}c_t + \mathbb{E}m_t)
	+ \mathbb{E}\sum_{j=1}^K \sum_{s\leq t, E_s^j \cap \mathcal{E}_{n_{s-1}^j}^{exp}} \Delta_{k_s}
= \Delta_{\max}(\mathbb{E}c_t + \mathbb{E}m_t)
	+ \sum_{j=1}^K \mathbb{E}R_t^{(4j)}
\: .\label{eq:regret_decomposition_with_easy_terms_bounded}
\end{align}

\subsection{Bounding $R_t^{(4j)}$}

$R_t^{(4j)}$ is the regret incurred when the algorithm explores and chooses the alternative set $\neg j$, and when the concentration event $\mathcal E^{exp}_{n_{t-1}^{j_t}}$ holds.

\subsubsection{Bounding $R_t^{(4*)}$}

For the special case of $*$, there is no regret when $n_{s-1}^* \text{ mod } 2 = 0$ (event $E_t^*\setminus E_t^{*\%2}$). Hence
$
R_t^{(4*)}
= \sum_{s\leq t, E_s^{*\%2} \cap \mathcal{E}_{n_{s-1}^*}^{exp}} \Delta^{k_s} \: .
$

\paragraph{Using the exploration criterion.}
Let $t+1$ be a stage in which there is exploration (event $E_{t+1}$). Remember that by hypothesis, all divergences are bounded by $C$. Then, using the tracking Theorem~\ref{thm:tracking}, $\vN_t \ge \sum_{s=1}^t \w_s - \log(K)$ and
\begin{align*}
f(t)
\geq \inf_{\vlambda \in \neg *} \sum_{k=1}^K N_t^k d(\hat{\theta}_t^k,  \lambda^k)
&\geq \inf_{\vlambda \in \neg *} \sum_{k=1}^K \sum_{s=1}^t w_s^k d(\hat{\theta}_t^k,  \lambda^k) - CK\log K\\
&= \inf_{\vlambda \in \neg *} \sum_{s=1}^t \sum_{k=1}^K w_s^k d(\hat{\theta}_t^k,  \lambda^k) - CK\log K
\: .
\end{align*}

\paragraph{Concentration to go to an online learning formulation.}
\begin{align*}
\inf_{\vlambda \in \neg *} \sum_{s=1}^t \sum_{k=1}^K w_s^k d(\hat{\theta}_t^k,  \lambda^k)
&\geq \inf_{\vlambda \in \neg *} \sum_{s\leq t, E_s^*} \sum_{k=1}^K w_s^k d(\hat{\theta}_t^k,  \lambda^k)\\
&=   \inf_{\vlambda \in \neg *} \sum_{s\leq t, E_s^*} \sum_{k=1}^K w_s^k d(\hat{\theta}_{s-1}^k,  \lambda^k)
	- \sum_{s\leq t, E_s^*} \sum_{k=1}^K w_s^k [d(\hat{\theta}_{s-1}^k,  \lambda^k) - d(\hat{\theta}_t^k,  \lambda^k)]
	\: .
\end{align*}
For all $\vlambda\in \neg *$, if $\mathcal{E}_{n_{s-1}^*}^{exp}$ then for all $k\in[K]$, $d(\hat{\theta}_{s-1}^k,  \lambda^k) - d(\hat{\theta}_t^k,  \lambda^k) \leq 2L\sqrt{2\sigma^2\frac{g_{s-1}(n_{s-1}^*)}{N_{s-1}^k}}$ by the Lipschitz assumption. If concentration does not hold (which happens for $m_t^*$ steps), it is smaller than $C$.
For $j\in[K]$, let $c_0^j(t) = \sum_{s\leq t, E_s^j\cap \mathcal{E}_{n_{s-1}^j}^{exp}} \sum_{k=1}^K w_s^k \sqrt{\frac{g_{s-1}(n_{s-1}^j)}{N_{s-1}^{k_s}}}$ . This quantity, of order $\sqrt{K n_t^j \log n_t^j}$, will appear several times in the following computations.
We proved
\begin{align*}
\inf_{\vlambda \in \neg *} \sum_{s=1}^t \sum_{k=1}^K w_s^k d(\hat{\theta}_t^k,  \lambda^k)
&\geq \inf_{\vlambda \in \neg *} \sum_{s\leq t, E_s^*} \sum_{k=1}^K w_s^k d(\hat{\theta}_{s-1}^k,  \lambda^k) - 2L c_0^*(t) - C m_t^*\\
&\geq \inf_{\vlambda \in \neg *} \sum_{s\leq t, E_s^{*\%2}} \sum_{k=1}^K w_s^k d(\hat{\theta}_{s-1}^k,  \lambda^k) - 2L c_0^*(t) - C m_t^*
\: .
\end{align*}

\paragraph{The $\vlambda$-player.}

By the regret property of the $\vlambda$-player algorithm,
\begin{align*}
\inf_{\vlambda \in \neg *} \sum_{s\leq t, E_s^{*\%2}} \sum_{k=1}^K w_s^k d(\hat{\theta}_{s-1}^k,  \lambda^k)
&\geq \sum_{s\leq t, E_s^{*\%2}} \sum_{k=1}^K w_s^k \ex_{\vlambda\sim\q_s^*}d(\hat{\theta}_{s-1}^k,  \lambda^k) - R_{n_t^*/2}^\vlambda
\: .
\end{align*}
Let $C_2(t) = 2L c_0^*(t) + C m_t^* + CK\log K + R_{n_t^*/2}^\lambda$. We now drop the rounds in which the concentration event does not hold from the sum.
\begin{align*}
f(t)
&\geq \sum_{s\leq t, E_s^{*\%2} \cap\mathcal{E}_{n_{s-1}^*}^{exp}} \sum_{k=1}^K w_s^k \ex_{\vlambda\sim\q_s^*} d(\hat{\theta}_{s-1}^k,  \lambda^k) - C_2(t)
\: .
\end{align*}

\paragraph{UCBs.}

We introduce the gap estimates $\tilde{\Delta}_s^k$ discussed in Section~\ref{sec:gap_estimates}.
\begin{align*}
f(t)
&\geq 
\sum_{s\leq t, E_s^{*\%2} \cap\mathcal{E}_{n_{s-1}^*}^{exp}} \sum_{k=1}^K w_s^k \ex_{\vlambda\sim\q_s} d(\hat{\theta}_{s-1}^k,  \lambda^k)
- C_2(t)
&=   \sum_{s\leq t, E_s^{*\%2} \cap\mathcal{E}_{n_{s-1}^*}^{exp}} \sum_{k=1}^K w_s^k \tilde{\Delta}_{s}^k \frac{\ex_{\vlambda\sim\q_s} d(\hat{\theta}_{s-1}^k,  \lambda^k)}{\tilde{\Delta}_{s}^k}
- C_2(t)
\: .
\end{align*}

We now introduce upper confidence bounds for the ratios in the expression above. We use $\UCB_s^k$ as discussed in Section~\ref{sec:ucb}.
\begin{align*}
f(t)
\geq \sum_{s\leq t, E_s^{*\%2} \cap\mathcal{E}_{n_{s-1}^*}^{exp}} \sum_{k=1}^K w_s^k
	\tilde{\Delta}_{s}^k \UCB_s^k
	- \sum_{s\leq t, E_s^{*\%2} \cap\mathcal{E}_{n_{s-1}^*}^{exp}} \sum_{k=1}^K w_s^k
		\tilde{\Delta}_{s}^k\left[ \UCB_s^k - \frac{\ex_{\vlambda\sim\q_s} d(\hat{\theta}_{s-1}^k,  \lambda^k)}{\tilde{\Delta}_{s}^k}\right]
	- C_2(t)
	\: .
\end{align*}
		
By the upper bound on $\UCB_s^k$ of equation~\eqref{eq:ucb_upper_bound}, and the Lipschitz property of $d$, for all $j\in[K]$,
\begin{align*}
&\sum_{s\leq t, E_s^{j\%2} \cap\mathcal{E}_{n_{s-1}^j}^{exp}} \sum_{k=1}^K w_s^k
		\tilde{\Delta}_{s}^k\left[ \UCB_s^k - \frac{\ex_{\vlambda\sim\q_s} d(\hat{\theta}_{s-1}^k,  \lambda^k)}{\tilde{\Delta}_s^k}\right]
\\
&\leq \sum_{s\leq t, E_s^{j\%2}\cap\mathcal{E}_{n_{s-1}^j}^{exp}} \sum_{k=1}^K w_s^k
	\left[ \sup_{\xi\in[\underline{\theta_{s-1}^k},\overline{\theta_{s-1}^k}]} \ex_{\vlambda\sim\q_s} d(\xi,  \lambda^k) - \ex_{\vlambda\sim\q_s} d(\hat{\theta}_{s-1}^k,  \lambda^k) \right]
\\
&\leq L \sum_{s\leq t, E_s^{j\%2}\cap\mathcal{E}_{n_{s-1}^j}^{exp}} \sum_{k=1}^K w_s^k
	\sup_{\xi\in[\underline{\theta_{s-1}^k},\overline{\theta_{s-1}^k}]} |\xi - \hat{\theta}_{s-1}^k|\\
&\leq L c_0^j(t)
\: .
\end{align*}

\paragraph{Regret of the $k$-player.}
Let $C_3(t) = C_2(t) + L c_0^*(t)$ .
We introduce $\tilde{w}_s^k = \frac{w_s^k \tilde{\Delta}_s^k}{\sum_{k=1}^K w_s^k \tilde{\Delta}_s^k}$, the proportion of ($\epsilon$-perturbed) regret allocated to arm $k$.
\begin{align*}
f(t)
&\geq \sum_{s\leq t, E_s^{*\%2} \cap\mathcal{E}_{n_{s-1}^*}^{exp}} \sum_{k=1}^K w_s^k
	\tilde{\Delta}_s^k \UCB_s^k
	- C_3(t)
= \sum_{s\leq t, E_s^{*\%2} \cap\mathcal{E}_{n_{s-1}^*}^{exp}} (\sum_{k=1}^K w_s^k \tilde{\Delta}_s^k)\sum_{k=1}^K \tilde{w}_s^k \UCB_s^k
	- C_3(t)
\: .
\end{align*}
We now use a regret-minimizing learner for that $\tilde{\w}$ action, where the loss of vector $\v\in \triangle_K$ at time $s$ is $\ell_s(\v)= -(\sum_{k=1}^K w_s^k \tilde{\Delta}_s^k)\sum_{k=1}^K v^k \UCB_s^k$.
Let $R_{n_t^*/2}^w$ be the regret of that learner after $n_t^*/2$ steps, such that
\begin{align*}
f(t)&\geq \max_{i\in[K]} \sum_{s\leq t, E_s^{*\%2} \cap\mathcal{E}_{n_{s-1}^*}^{exp}}
	\left(\sum_{k=1}^K w_s^k \tilde{\Delta}_s^k\right)  \UCB_s^i
	- R_{n_t^*/2}^w - C_3(t)
\: .
\end{align*}

\paragraph{Using optimism.}

We use the lower bound on $\UCB_s^i$ of equation~\eqref{eq:ucb_lower_bound}. Under $\mathcal{E}^{exp}_{n_{s-1}^*}$, with $\Delta_s^i = \max\{\varepsilon_s, \Delta^i\}$,
\begin{align*}
\UCB_s^i
\geq \frac{\ex_{\vlambda\sim \q_s} d(\theta^i, \lambda^i)}{\Delta_s^i}(1 - \sqrt{\frac{8\sigma^2 g_{s-1}( n_{s-1}^*)}{(\Delta_s^i)^2 N_{s-1}^*}})
\end{align*}
Let $c_4(t) = \max_{i\neq *}\sum_{s\leq t, E_s^{*\%2} \cap\mathcal{E}_{n_{s-1}^*}^{exp}}
(\sum_{k=1}^K w_s^k \tilde{\Delta}_s^k) \frac{\ex_{\vlambda\sim \q_s} d(\theta^i, \lambda^i)}{\Delta_s^i}\sqrt{\frac{8\sigma^2 g_{s-1}(n_{s-1}^*)}{(\Delta_s^i)^2 N_{s-1}^*}}$ and let $C_4(t) = C_3(t) + R_{n_t^*/2}^w + c_4(t)$. Then
\begin{align}
f(t)
&\geq \max_{i\in[K]} \sum_{s\leq t, E_s^{*\%2} \cap\mathcal{E}_{n_{s-1}^*}^{exp}}
	\left(\sum_{k=1}^K w_s^k \tilde{\Delta}_s^k\right) \frac{\ex_{\vlambda\sim \q_s} d(\theta^i, \lambda^i)}{\Delta_s^i}
	- C_4(t)
\: .\label{eq:before_removing_first_epsilon}
\end{align}

\paragraph{An inequality on the $\varepsilon$-perturbed regret.}

For $s\in[t]$, let $\alpha_s = \frac{\sum_{k=1}^K w_s^k \tilde{\Delta}_s^k}{\sum_{s\leq t, E_s^{*\%2} \cap\mathcal{E}_{n_{s-1}^*}^{exp}}
\sum_{k=1}^K w_s^k \tilde{\Delta}_s^k}$. These are positive and sum to 1.
\begin{align*}
f(t)
&\geq \left( \sum_{s\leq t, E_s^{*\%2}\cap\mathcal{E}_{n_{s-1}^*}^{exp}}
\sum_{k=1}^K w_s^k \tilde{\Delta}_s^k \right)
\max_k \sum_{s\leq t, E_s^{*\%2} \cap\mathcal{E}_{n_{s-1}^*}^{exp}}
\alpha_s \frac{\ex_{\vlambda\sim \q_s} d(\theta^k, \lambda^k)}{\Delta_s^k}
- C_4(t)
\: .
\end{align*}
We use that $\tilde{\Delta}_{s}^k \geq \Delta_s^k - \indicator\{k\neq *\} 2\sqrt{2\sigma^2\frac{g_{s-1}(n_{s-1}^*)}{N_{s-1}^k}}$. Remark that 
$2\sum_{s\leq t, E_s^{*\%2} \cap\mathcal{E}_{n_{s-1}^*}^{exp}, k_s\neq *} \sum_{k=1}^K w_s^k
\sqrt{2\sigma^2\frac{g_{s-1}(n_{s-1}^*)}{N_{s-1}^k}} \leq 2 c_0^*(t)$.
\begin{align*}
f(t)
&\geq \left( \sum_{s\leq t, E_s^{*\%2}\cap\mathcal{E}_{n_{s-1}^*}^{exp}} \sum_{k=1}^K w_s^k \Delta_s^k - 2 c_0^*(t)\right)
\max_k \sum_{s\leq t, E_s^{*\%2} \cap\mathcal{E}_{n_{s-1}^*}^{exp}}
\alpha_s \frac{\ex_{\vlambda\sim \q_s} d(\theta^k, \lambda^k)}{\Delta_s^k}
- C_4(t)
\: .
\end{align*}

If all $(\varepsilon_s)_{s\ge 1}$ are equal, let $\Delta^k_\varepsilon = \max\{\varepsilon, \Delta^k\}$. The sum of expectations $\sum_{s} \alpha_s \ex_{\q_s}$ is the expectation under another distribution over $\vlambda$. Hence
\begin{align*}
f(t)
\geq
\left( \sum_{s\leq t, E_s^{*\%2}\cap\mathcal{E}_{n_{s-1}^*}^{exp}} \sum_{k=1}^K w_s^k \Delta_\varepsilon^k - 2 c_0^*(t) \right)
\inf_{\q} \max_k \frac{\ex_{\vlambda\sim \q} d(\theta^k, \lambda^k)}{\Delta_\varepsilon^k}
- C_4(t)
\: .
\end{align*}
Let $D_\varepsilon = \inf_{\q} \max_k \frac{\ex_{\vlambda\sim \q} d(\theta^k, \lambda^k)}{\Delta_\varepsilon^k}$. This is smaller than $D = \inf_{\q} \max_k \frac{\ex_{\vlambda\sim \q} d(\theta^k, \lambda^k)}{\Delta^k}$.
Let $C_5(t) = \frac{1}{D}C_4(t) + 2 c_0^*(t)$.
The bound on the pertubed regret (perturbed since function of the gaps $\Delta_\varepsilon^k$) is
\begin{align*}
\frac{D_\varepsilon}{D}\sum_{s\leq t, E_s^{*\%2}\cap\mathcal{E}_{n_{s-1}^*}^{exp}} \sum_{k=1}^K w_s^k \Delta_\varepsilon^k
\leq \frac{f(t)}{D} + C_5(t)
\: .
\end{align*}

In general, if $(\varepsilon_s)_{s\ge 1, j_s=*}$ is a non-increasing sequence, we remove the first terms of the sum in~\eqref{eq:before_removing_first_epsilon},
\begin{align*}
f(t)
&\geq \max_{i\in[K]} \sum_{s\leq t, n_s^* \ge \sqrt{n_t^*}, E_s^{*\%2} \cap\mathcal{E}_{n_{s-1}^*}^{exp}}
	\left(\sum_{k=1}^K w_s^k \tilde{\Delta}_s^k\right) \frac{\ex_{\vlambda\sim \q_s} d(\theta^i, \lambda^i)}{\Delta_s^i}
	- C_4(t)
\: .
\end{align*}
We introduce positive coefficients $\alpha$ that sum to 1 as above and follow the same steps, such that
\begin{align*}
f(t)
&\geq \left( \sum_{ s\leq t, n_s^* \ge \sqrt{n_t^*}, E_s^{*\%2}\cap\mathcal{E}_{n_{s-1}^*}^{exp}} \sum_{k=1}^K w_s^k \Delta_s^k - 2 c_0^*(t)\right)
\max_k \sum_{s\leq t, n_s^* \ge \sqrt{n_t^*}, E_s^{*\%2} \cap\mathcal{E}_{n_{s-1}^*}^{exp}}
\alpha_s \frac{\ex_{\vlambda\sim \q_s} d(\theta^k, \lambda^k)}{\Delta_s^k}
- C_4(t)
\\
&\ge  \left( \sum_{ s\leq t, E_s^{*\%2}\cap\mathcal{E}_{n_{s-1}^*}^{exp}} \sum_{k=1}^K w_s^k \Delta_s^k - \max\{\varepsilon_0,\Delta_{\max}\}\sqrt{n_t^*} - 2 c_0^*(t)\right)
\max_k \sum_{s\leq t, n_s^* \ge \sqrt{n_t^*}, E_s^{*\%2} \cap\mathcal{E}_{n_{s-1}^*}^{exp}}
\alpha_s \frac{\ex_{\vlambda\sim \q_s} d(\theta^k, \lambda^k)}{\Delta_s^k}
- C_4(t)
\: .
\end{align*}
Now let $\varepsilon(n_t^*)$ be the maximum of all $\varepsilon_s$ for $s$ such that $s\leq t, n_s^* \ge \sqrt{n_t^*}, E_s^{*\%2} \cap\mathcal{E}_{n_{s-1}^*}$. Then
\begin{align*}
f(t)
&\ge  \left( \sum_{ s\leq t, E_s^{*\%2}\cap\mathcal{E}_{n_{s-1}^*}^{exp}} \sum_{k=1}^K w_s^k \Delta_s^k - \max\{\varepsilon_0,\Delta_{\max}\}\sqrt{n_t^*} - 2 c_0^*(t)\right)
\inf_{\q} \max_k \frac{\ex_{\vlambda\sim \q_s} d(\theta^k, \lambda^k)}{\max\{\varepsilon(n_t^*),\Delta^k\}}
- C_4(t)
\: .
\end{align*}
We have
\begin{align*}
\frac{D_{\varepsilon(n_t^*)}}{D}\sum_{s\leq t, E_s^{*\%2}\cap\mathcal{E}_{n_{s-1}^*}^{exp}} \sum_{k=1}^K w_s^k \Delta_s^k
\leq \frac{f(t)}{D} + C_5(t) + \max\{\varepsilon_0,\Delta_{\max}\}\sqrt{n_t^*}
\: .
\end{align*}

By Theorem~\ref{thm:perturbation.cost}, there exists $c>0$ such that $D_{\varepsilon(n_t^*)} \ge D - c\sqrt{\varepsilon(n_t^*) D}$, hence
\begin{align*}
\left(1 - c \sqrt{\frac{\varepsilon(n_t^*)}{D}}\right)\sum_{s\leq t, E_s^{*\%2}\cap\mathcal{E}_{n_{s-1}^*}^{exp}} \sum_{k=1}^K w_s^k \Delta_s^k
\leq \frac{f(t)}{D} + C_5(t) + \max\{\varepsilon_0,\Delta_{\max}\}\sqrt{n_t^*}
\: .
\end{align*}

\paragraph{Inequality on the regret.}
Let $R_{t,\varepsilon}^{(4*)}$ be the perturbed regret. Then with $\varepsilon_t^*$ the smallest $\varepsilon_s$ for $s\le t$ such that $E_s^{*\%2}\cap\mathcal{E}_{n_{s-1}^*}$,
\begin{align*}
\varepsilon_t^* (n_t^*/2 - m_t^*)
\le \sum_{s\leq t, E_s^{*\%2}\cap\mathcal{E}_{n_{s-1}^*}^{exp}} \varepsilon_s
\le \sum_{s\leq t, E_s^{*\%2}\cap\mathcal{E}_{n_{s-1}^*}^{exp}} \sum_{k=1}^K w_s^k \Delta_s^k
= R_{t,\varepsilon}^{(4*)}
\: .
\end{align*}

From Lemma~\ref{lem:computation_of_all_c} and with $H$ as defined in that Lemma, $C_5(t) \le H \sqrt{n_t^* g_t(n_t^*)} + \frac{1}{D}\left[R_{n_t^*/2}^w + R_{n_t^*/2}^\lambda + Cm_t^*\right]$, unless $n_t^*<K$ or $g_t(n_t^*)>n_t^*/\log^2(n_t^*)$.

If $n_t^* < K$ then $R_t^{(4*)} < K \Delta_{\max}$. If $g_t(n_t^*)>n_t^*/\log^2(n_t^*)$ then there exists a constant $G$ such that $n_t^* \le G\log\log t$ and the regret is again bounded by that number times $\Delta_{\max}$. If none of these two inequalities are true, let $R$ be such that $R_{n_t^*/2}^w + R_{n_t^*/2}^\lambda \le R \sqrt{n_t^*}$. We have
\begin{align*}
\left(1 - c \sqrt{\frac{\varepsilon(n_t^*)}{D}}\right)R_{t,\varepsilon}^{(4*)}
&\leq \frac{f(t)}{D}
+ H \sqrt{n_t^* g_t(n_t^*)} + \frac{1}{D}\left[R\sqrt{n_t^*} + Cm_t^*\right]
\\
&\leq \frac{f(t)}{D}
+ (H+\frac{R}{D}) \sqrt{(2\frac{R_{t,\varepsilon}^{(4*)}}{\varepsilon_t^*}+m_t^*) g_t(2\frac{R_{t,\varepsilon}^{(4*)}}{\varepsilon_t^*}+m_t^*)} + \frac{C}{D}m_t^*
\: .
\end{align*}

Let $y = 2\frac{R_{t,\varepsilon}^{(4*)}}{\varepsilon_t^*}+m_t^*$, such that for a constant $\varepsilon$ sequence, the equation above implies
\begin{align*}
\left(1 - c \sqrt{\frac{\varepsilon}{D}}\right)y \le \frac{2f(t)}{ \varepsilon D}
+ \frac{2}{\varepsilon}\left(H+\frac{R}{D}\right) \sqrt{y g_t(y)} + \left(\frac{2C}{ \varepsilon D} +1\right)m_t^*
\: .
\end{align*}

We need to solve an equation of the form $y \le A + B\sqrt{y g_t(y)}$. By concavity of $y\mapsto \sqrt{y g_t(y)}$ (which comes from the concavity of $y\mapsto \sqrt{y \log(y)}$), we have the same inequality in expectation: $\mathbb{E}y \le \mathbb{E}A + B\sqrt{\mathbb{E}y g_t(\mathbb{E}y)}$.

For $A \gg B$ we get approximately $\mathbb{E}y \le A$, i.e. $\mathbb{E} R_{t,\varepsilon}^{(4*)} \le \frac{f(t) + C \mathbb{E}m_t^*}{D(1 - c \sqrt{\frac{\varepsilon}{D}})} \approx \frac{f(t)}{D(1 - c \sqrt{\frac{\varepsilon}{D}})}$ since $\mathbb{E}m_t^*$ is of order $\log\log t$ and $f(t) \approx \log(t)$.

For $\varepsilon_s = 1/(n_s^{j_s})^{1/a}$ with $a$ big enough, we have by construction $\varepsilon(n_t^*) \le 1/(n_t^{*})^{1/(2a)}$, $\varepsilon_t^* \ge 1/(n_t^{*})^{1/a}$ and we get an equation on the regret that implies asymptotic optimality.

\subsubsection{Bounding $R_t^{(4j)}$ for $j\neq *$}\label{sec:misplore}

We prove an upper bound on $n_t^j$. Then $R_t^{(4j)} \leq \Delta_{\max} n_t^j$.
We first show that there exists $k$ for which $\UCB_s^k$ is bigger than some constant.

\begin{lemma}
For all $s\le t$ such that $j_s \ne i^*(\theta)$ and the event $\mathcal{E}_{n_{s-1}^{j_s}}^{exp}$ holds, there exists an arm $k\in [K]$ such that
\begin{align*}
\UCB_s^k
&\geq \frac{\Delta_{\min}^2}{8\sigma^2} \frac{1}{\max\left\{ \varepsilon_s, \Delta_{\max} + \Delta_{\min} + 2\sqrt{2 \sigma^2 g_t(n_{s-1}^{j_s})}  \right\} }
\: .
\end{align*}
\end{lemma}
\begin{proof}
Set a time $s$ and let $j=j_s \neq i^*(\theta)$.
Since $i^*(\hat{\theta}_{s-1}) \neq i^*(\theta)$, there exists $k$ such that $|\hat{\theta}_{s-1}^k - \theta^k| > \Delta_{\min}/2$ .
Then by the sub-Gaussian property of the arm distributions, for that arm,
\begin{align*}
d(\theta^k, \hat{\theta}_{s-1}^k)
\ge \frac{1}{2 \sigma^2}(\hat{\theta}_{s-1}^k - \theta^k)^2
\ge \frac{\Delta_{\min}^2}{8 \sigma^2}
\: .
\end{align*}
Under $\mathcal{E}_{n_{s-1}^j}^{exp}$,
\begin{align*}
\max_{\xi \in [\underline{\theta_{s-1}^k}, \overline{\theta_{s-1}^k}]}\ex_{\vlambda\sim\q_s}d(\xi, \lambda^k)
\ge \max_{\xi \in [\underline{\theta_{s-1}^k}, \overline{\theta_{s-1}^k}]} d(\xi, \hat{\theta}_{s-1}^k)
\ge d(\theta^k, \hat{\theta}_{s-1}^k)
\: .
\end{align*}

We obtain that for the arm $k$ discussed above,
\begin{align*}
\UCB_s^k
= \max_{\xi\in[\underline{\theta_{s-1}^k}, \overline{\theta_{s-1}^k}]} \left[ \ex_{\vlambda\sim\q_s}\frac{d(\xi, \lambda^k)}{\max\{\varepsilon_s, \overline{\theta^{j}_{s-1}} - \xi\}} \right]
&\geq  \frac{\max_{\xi \in [\underline{\theta^{k}_{s-1}},\overline{\theta^{k}_{s-1}}]}\ex_{\vlambda\sim\q_s}d(\xi, \lambda^k)}
{\max\{\varepsilon_s, \overline{\theta^{j}_{s-1}} - \underline{\theta^{k}_{s-1}}\}}
\\
&=  \frac{\max_{\xi \in [\underline{\theta^{k}_{s-1}},\overline{\theta^{k}_{s-1}}]}\ex_{\vlambda\sim\q_s}d(\xi, \lambda^k)}
{\max\{\varepsilon_s, \theta^j  - \theta^k + (\overline{\theta^{j}_{s-1}}- \theta^j) + (\theta^k - \underline{\theta^{k}_{s-1}})\}}
\: .
\end{align*}
Under $\mathcal{E}_{n_{s-1}^j}^{exp}$,
\begin{align*}
\UCB_s^k
&\geq \frac{\max_{\xi \in [\underline{\theta^{k}_{s-1}},\overline{\theta^{k}_{s-1}}]}\ex_{\vlambda\sim\q_s}d(\xi, \lambda^k)}
{\max\{\varepsilon_s, \theta^j  - \theta^k + (\overline{\theta^{j}_{s-1}}- \underline{\theta_{s-1}^j}) + (\overline{\theta_{s-1}^k} - \underline{\theta^{k}_{s-1}})\}}
\\
&\geq \frac{\Delta_{\min}}{8 \sigma^2}
\frac{1}{\max\{\varepsilon_s, \theta^j  - \theta^k + (\overline{\theta^{j}_{s-1}}- \underline{\theta_{s-1}^j}) + (\overline{\theta_{s-1}^k} - \underline{\theta^{k}_{s-1}})\}}
\: .
\end{align*}
Then if $\overline{\theta^{j}_{s-1}}- \underline{\theta_{s-1}^j} \leq \Delta_{\min}$ , under $\mathcal{E}_{n_{s-1}^j}^{exp}$,
\begin{align*}
\UCB_s^k
&\geq \frac{\Delta_{\min}^2}{8\sigma^2} \frac{1}{\max\left\{\varepsilon_s, \theta^j - \theta^k + \Delta_{\min} + 2\sqrt{ 2 \sigma^2\frac{g_{s-1}(n_{s-1}^j)}{N_{s-1}^k}} \right\} }
\\
&\geq \frac{\Delta_{\min}^2}{8\sigma^2} \frac{1}{\max\left\{ \varepsilon_s, \Delta_{\max} + \Delta_{\min} + 2\sqrt{2 \sigma^2 g_t(n_{s-1}^j)}  \right\} }
\: .
\end{align*}

If $\overline{\theta^{j}_{s-1}}- \underline{\theta_{s-1}^j} > \Delta_{\min}$, then
\begin{align*}
\max_{\xi \in [\underline{\theta^{j}_{s-1}}, \overline{\theta^{j}_{s-1}}]} d(\xi, \hat{\theta}_{s-1}^j)
\ge \frac{1}{2 \sigma^2}\max_{\xi \in [\underline{\theta^{j}_{s-1}},\overline{\theta^{j}_{s-1}}]} (\xi - \hat{\theta}_{s-1}^j)^2
\ge \frac{1}{8 \sigma^2} (\overline{\theta_{s-1}^j} - \underline{\theta_{s-1}^j})^2
&\ge \frac{\Delta_{\min}^2}{8 \sigma^2} \: ,
\end{align*}
and we get that $\UCB_s^j \ge \frac{\Delta_{\min}^2}{8 \sigma^2 \varepsilon_s}$.

\end{proof}

Suppose that $(\varepsilon_s)_{s\in\N, j_s=j}$ is non-increasing. 
 We proved that for all $s$ such that $E_s^j\cap\mathcal{E}_{n_{s-1}^j}^{exp}$, there exists $k\in[K]$ with $(\sum_{k=1}^K w_s^k \tilde{\Delta}_s^k)\UCB_s^k \geq (\sum_{k=1}^K w_s^k \tilde{\Delta}_s^k)\frac{\Delta_{\min}^2}{16\sigma^2(\max\{\varepsilon_0,\Delta_{\max}\}+ \sqrt{2 \sigma^2 g_t(n_t^j)})}$. Let $\Delta_{\max,0} = \max\{\varepsilon_0,\Delta_{\max}\}$. Then there exists also a $k'\in[K]$ for which
\begin{align*}
\sum_{s\le t, E_s^j\cap\mathcal{E}_{n_{s-1}^j}^{exp}}\left(\sum_{k=1}^K w_s^k \tilde{\Delta}_s^k\right)\UCB_s^{k'}
\ge \frac{1}{K} \sum_{s\le t, E_s^j\cap\mathcal{E}_{n_{s-1}^j}^{exp}}\left(\sum_{k=1}^K w_s^k \tilde{\Delta}_s^k\right)\frac{\Delta_{\min}^2}{16\sigma^2\left(\Delta_{\max,0}+ \sqrt{2 \sigma^2 g_t(n_t^j)}\right)}
\: .
\end{align*}

Since $\vtheta \in \neg j$,
\begin{align*}
\sum_{s\leq t, E_s^j \cap \mathcal{E}_{n_{s-1}^j}^{exp}} \sum_{k=1}^K w_s^k d(\hat{\theta}_{s-1}^k, \theta^k)
&\geq \inf_{\vlambda \in \neg j}\sum_{s\leq t, E_s^j \cap \mathcal{E}_{n_{s-1}^j}^{exp}} \sum_{k=1}^K w_s^k d(\hat{\theta}_{s-1}^k, \lambda^k)
\: .
\end{align*}

Left hand side:
\begin{align*}
\sum_{s\leq t, E_s^j \cap \mathcal{E}_{n_{s-1}^j}^{exp}} \sum_{k=1}^K w_s^k d(\hat{\theta}_{s-1}^k, \theta^k)
\leq \sum_{s\leq t, E_s^j \cap \mathcal{E}_{n_{s-1}^j}^{exp}} \sum_{k=1}^K \frac{w_s^k}{N_{s-1}^k} g_{s-1}(n_{s-1}^j)
\le  2K \log(n_t^j)g_t(n_t^j)
\: .
\end{align*}

Right hand side: we apply the same proof steps as for the bounding of $R_t^{(4*)}$.
\begin{align*}
\inf_{\vlambda \in \neg j}\sum_{s\leq t, E_s^j \cap \mathcal{E}_{n_{s-1}^j}^{exp}} \sum_{k=1}^K w_s^k d(\hat{\theta}_{s-1}^{k_s}, \lambda^{k_s})
&\geq \inf_{\vlambda \in \neg j}\sum_{s\leq t, E_s^{j\%2} \cap \mathcal{E}_{n_{s-1}^j}^{exp}} \sum_{k=1}^K w_s^k d(\hat{\theta}_{s-1}^{k_s}, \lambda^{k_s})\\
&\geq \inf_{\vlambda \in \neg j}\sum_{s\leq t, E_s^{j\%2}} \sum_{k=1}^K w_s^k d(\hat{\theta}_{s-1}^{k_s}, \lambda^{k_s}) - C m_t^j\\
&\geq \sum_{s\leq t, E_s^{j\%2}} \sum_{k=1}^K w_s^k \ex_{\vlambda\sim\q_s} d(\hat{\theta}_{s-1}^{k_s}, \lambda^{k_s}) - C m_t^j - R_{n_t^j}^\lambda
\: .
\end{align*}

\begin{align*}
\sum_{s\leq t, E_s^{j\%2}} \sum_{k=1}^K w_s^k \ex_{\vlambda\sim\q_s} d(\hat{\theta}_{s-1}^{k_s}, \lambda^{k_s})
&\geq \sum_{s\leq t, E_s^{j\%2} \cap \mathcal{E}_{n_{s-1}^j}^{exp}} \sum_{k=1}^K w_s^k \tilde{\Delta}_s^k \ex_{\vlambda\sim\q_s} \frac{d(\hat{\theta}_{s-1}^k, \lambda^k)}{\tilde{\Delta}_s^k}
\\
&\geq \sum_{s\leq t, E_s^{j\%2} \cap \mathcal{E}_{n_{s-1}^j}^{exp}} \sum_{k=1}^K w_s^k \tilde{\Delta}_s^k \UCB_s^k
- L c_0^j(t)
\\
&\geq \max_{i\in[K]} \sum_{s\leq t, E_s^{j\%2} \cap \mathcal{E}_{n_{s-1}^j}^{exp}} \left(\sum_{k=1}^K w_s^k \tilde{\Delta}_s^k\right) \UCB_s^i
- R_{n_t^j/2}^w - L c_0^j(t)
\: .
\end{align*}

As we argued above, that maximum is such that
\begin{align*}
\max_{i\in[K]} \sum_{s\leq t, E_s^{j\%2} \cap \mathcal{E}_{n_{s-1}^j}^{exp}} \left(\sum_{k=1}^K w_s^k \tilde{\Delta}_s^k\right) \UCB_s^i
&\ge \frac{1}{K} \sum_{s\leq t, E_s^{j\%2} \cap \mathcal{E}_{n_{s-1}^j}^{exp} }
\left(\sum_{k=1}^K w_s^k \tilde{\Delta}_s^k\right) \frac{\Delta_{\min}^2}{16\sigma^2\left(\Delta_{\max,0}+ \sqrt{2 \sigma^2 g_t(n_t^j)}\right)}
\\
&\ge (n_t^j/2 - m_t^j) \frac{1}{K}\varepsilon_t^j\frac{\Delta_{\min}^2}{16\sigma^2\left(\Delta_{\max,0}+ \sqrt{2 \sigma^2 g_t(n_t^j)}\right)}
\: .
\end{align*}
In the last line, $\varepsilon_t^j = \varepsilon_{t'}$ where $t'$ is the last time before $t$ with $j_t = j$.

We obtain that $n_t^j$ verifies
\begin{align*}
2K \log(n_t^j)g_t(n_t^j)
\geq 
(n_t^j/2 - m_t^j) \frac{\varepsilon_t^j}{K}
\frac{\Delta_{\min}^2}{16\sigma^2\left(\Delta_{\max,0}+ \sqrt{2 \sigma^2 g_t(n_t^j)}\right)}
- R_{n_t^j/2}^w - R_{n_t^j/2}^\lambda - C m_t^j - L c_0^j(t)
\: .
\end{align*}

Let $R$ be such that $R_{n_t^j/2}^w + R_{n_t^j/2}^\lambda \le R \sqrt{n_t^j}$. From Lemma~\ref{lem:computation_of_all_c}, either $n_t^j \le K$ or $c_0^j(t) \le K (\log K+ \sqrt{8}) \sqrt{2 \sigma^2 n^j_t g_t(n^j_t)}$. Then either $n_t^j \le G \log\log t$ for a constant $G$ (that depends on $K$), or $\log(n_t^j)\sqrt{g_t(n_t^j)} \le \sqrt{n_t^j}$. When $n_t^j > \max\{K, G\log\log t\}$, we have
\begin{align*}
n_t^j/2
&\le m_t + 
	\frac{16K \sigma^2}{\Delta_{\min}^2 \varepsilon_t^j}
		\left[
		2K\Delta_{\max,0}\log(n_t^j)g_t(n_t^j) + R\Delta_{\max,0} \sqrt{n_t^j} + LK\Delta_{\max,0}(\log(K)+\sqrt{8})\sqrt{2 \sigma^2 n_t^j g_t(n_t^j)}
		\right.
		\\
		& \qquad \qquad \qquad \quad 
		+ \left. 2K\log(n_t^j)g_t(n_t^j)\sqrt{2 \sigma^2 g_t(n_t^j)} + R \sqrt{2 \sigma^2 n_t^j g_t(n_t^j)} + LK(\log(K)+\sqrt{8})2 \sigma^2 \sqrt{n_t^j} g_t(n_t^j)
		\right.
		\\
		& \qquad \qquad \qquad \quad 
		+ \left.
		C m_t\left(\Delta_{\max,0} + \sqrt{2 \sigma^2 g_t(n_t^j)}\right)
	\right]
\\
&\le m_t + 
	\frac{16K \sigma^2}{\Delta_{\min}^2 \varepsilon_t^j} \sqrt{n_t^j} g(n_t^j)
		\left[
		2K\Delta_{\max,0} + R\Delta_{\max,0} + LK\Delta_{\max,0}(\log(K)+\sqrt{8})\sqrt{2 \sigma^2}
		\right.
		\\
		& \qquad \qquad  \qquad \qquad \qquad \quad 
		+ \left. 2K\sqrt{2 \sigma^2}  + R \sqrt{2 \sigma^2} + LK(\log(K)+\sqrt{8})2 \sigma^2
		\right]
		\\
	& \qquad  \quad 
	+ \frac{16K \sigma^2}{\Delta_{\min}^2 \varepsilon_t^j}
	\left[
		C m_t\left(\Delta_{\max,0} + \sqrt{2 \sigma^2 g_t(n_t^j)}\right)
	\right]
	\: .
\end{align*}
Bounding the last $\sqrt{g_t(n_t^j)}$ by $\sqrt{g_t(t)}$, we obtain an equation of the form $\varepsilon_t^j n_t^j \le \alpha_1 +  \alpha_2 \sqrt{n_t^j}g_t(n_t^j) + m_t(\alpha_3+ \alpha_4\sqrt{g_t(t)})$. The quantities $\alpha_1,\alpha_2,\alpha_3,\alpha_4$ in that inequality are deterministic. By concavity of the function of $n_t^j$ present on the right hand side, we have
\begin{align*}
\mathbb{E} [\varepsilon_t^j n_t^j] \le \alpha_1 +  \alpha_2 \sqrt{\mathbb{E}n_t^j}g_t(\mathbb{E}n_t^j) + \mathbb{E}m_t(\alpha_3+ \alpha_4\sqrt{g_t(t)}) \: .
\end{align*}
If $\varepsilon_t^j$ is constant, it only remains to solve that inequality to get a bound of the form $\mathbb{E}n_t^j = \mathcal O(\sqrt{\log t})$ where the dominant term is $\sqrt{g_t(t)} \approx \sqrt{\log t}$. Hence $\mathbb{E}R_t^{(4j)} = \mathcal O(\sqrt{\log t})$ as well since $\mathbb{E}R_t^{(4j)} \le \Delta_{\max} n_t^j$.

If we take $\varepsilon_s = 1/(n_s^{j_s})^{1/a}$ for $a>2$, we also get a bound $\mathcal O((\log t)^{a/(2(a-1))})$ .

\subsection{Regret bound: summary} 
\label{sub:regret_bound_summary}

Recall that we proved~\eqref{eq:regret_decomposition_with_easy_terms_bounded} in Section~\ref{sub:easy_regret_terms}:
\begin{align*}
\mathbb{E} R_t
&\le \Delta_{\max}(\mathbb{E}c_t + \mathbb{E}m_t)
	+ \sum_{j=1}^K \mathbb{E}R_t^{(4j)}
\: .
\end{align*}
The two first terms are $\mathcal O (\log\log t)$. For the last one, we proved that for $j\neq *$, $\mathbb{E}R_t^{(4j)} = \mathcal O(\sqrt{\log t})$ and that $\mathbb{E}R_t^{(4*)}$ is asymptotically equivalent to $f(t)/D_{\varepsilon}$ (for fixed $\varepsilon$), with $f(t) \approx \log(t)$. Hence that ratio is also the dominant term of the whole regret. By Theorem~\ref{thm:perturbation.cost}, $D_{\varepsilon} \ge D - c\sqrt{\varepsilon D}$ for a constant $c$.

In order to get asymptotic optimality, we choose $\varepsilon_s = 1/(n_s^{j_s})^{1/a}$ for $a>2$ big enough.

It is of course possible to get a concrete, explicit bound for the regret from the results in this section. As we foster hardly any hope that the resulting lower order terms are either representative or necessary, we instead focused on the asymptotically dominant term (while providing some hints on the time at which it becomes dominant).


\section{Technical Lemmas}\label{app:technical}

\begin{lemma}
C-Tracking verifies
\begin{align*}
\sum_{s=1}^t \sum_{k=1}^K \frac{w_s^k}{\sqrt{N_{s-1}^k}}
&\le K\log K + 2\sqrt{2 K t} \: .
\\
\sum_{s=1}^t \sum_{k=1}^K \frac{w_s^k}{N_{s-1}^k}
&\le 2K \log(t) \: .
\end{align*}
\end{lemma}
The proof is the same as for Lemma 9 of \cite{degenne2019non}, except that a factor $K$ is replaced by $\log K$ due to our improved tracking result (see Theorem~\ref{thm:tracking}).

\begin{lemma}\label{lem:computation_of_all_c}
We have the inequalities
\begin{align*}
c_0^j(t)
&\le \sqrt{2\sigma^2 g_t(n_t^j)}(K\log K + 2\sqrt{2 K n_t^j}) \: ,
\\
c_4(t)
&\le 4C\frac{\max\{\varepsilon_1,\Delta_{\max}\}}{\Delta_{\min}^2} \sqrt{2\sigma^2 n_t^* g_t(n_t^*)} + \frac{8\sigma^2 C}{\Delta_{\min}} \log(n_t) g_t(n_t^*)
\: ,
\end{align*}
where these quantities are defined in the sample complexity proof. Furthermore, if $g(n_t^*) \le n_t^*/\log^2(n_t^*)$ and $n_t^*\ge K$,
\begin{align*}
c_0^*(t)
&\le \sqrt{K}(\log K+ \sqrt{8})\sqrt{2 \sigma^2 n_t^* g_t(n_t^*)} \: ,
\\
c_4(t)
&\le \frac{4C}{\Delta_{\min}}\left(\frac{\max\{\varepsilon_1,\Delta_{\max}\}}{\Delta_{\min}}  + \sqrt{2\sigma^2} \right)\sqrt{2\sigma^2 n_t^* g_t(n_t^*)}
\: ,
\\
C_5(t)
&\le \left( \frac{4C}{\Delta_{\min}D}(\frac{\max\{\varepsilon_1,\Delta_{\max}\}}{\Delta_{\min}}  + \sqrt{2\sigma^2} ) +  (2+\frac{3L}{D})\sqrt{K}(\log K+ \sqrt{8})\right) \sqrt{2 \sigma^2 n_t^* g_t(n_t^*)}
\\&\quad + \frac{1}{D}\left[R_{n_t^*/2}^w + R_{n_t^*/2}^\lambda + Cm_t^*\right] \: .
\end{align*}
We define $H$ as the term such that this last inequality reads $C_5(t) \le H \sqrt{n_t^* g_t(n_t^*)} + \frac{1}{D_\varepsilon}\left[R_{n_t^*/2}^w + R_{n_t^*/2}^\lambda + Cm_t^*\right]$ .
\end{lemma}

\begin{proof}
Let $c_0^j(t) = \sum_{s\leq t, E_s^j\cap \mathcal{E}_{n_{s-1}^j}^{exp}}\sum_{k=1}^k w_s^k \sqrt{\frac{2\sigma^2 g_{s-1}(n_{s-1}^j)}{N_{s-1}^k}}$. Then
\begin{align*}
c_0^j(t)
\le \sqrt{2\sigma^2 g_t(n_t^j)} \sum_{s\le t, E_s^j} \sum_{k=1}^K \frac{w_s^k}{\sqrt{N_{s-1}^k}}
\le \sqrt{2\sigma^2 g_t(n_t^j)}(K\log K + 2\sqrt{2 K n_t^j})
\: .
\end{align*}

We now bound $c_4(t)$.
\begin{align*}
c_4(t)
&= \max_{i\neq *}\sum_{s\leq t, E_s^{*\%2}\cap\mathcal{E}_{n_{s-1}^*}^{exp}} \sum_{k=1}^k w_s^k
\tilde{\Delta}_{s}^k \frac{\ex_{\vlambda\sim \q_s} d(\theta^i, \lambda^i)}{\Delta_s^i}\sqrt{\frac{8\sigma^2 g_{s-1}(n_{s-1}^*)}{(\Delta_s^i)^2 N_{s-1}^*}}
\: .
\end{align*}

By equation~\eqref{eq:gap_estimation_upper_bound},
\begin{align*}
\tilde{\Delta}_s^k
\leq \Delta_s^k + 2\sqrt{2\sigma^2\frac{g_{s-1}(n_t^*)}{N_{s-1}^*}} 
\leq \Delta_s^k + 2\sqrt{2\sigma^2\frac{g_{t}(n_t^*)}{N_{s-1}^*}}
\leq \max\{\varepsilon_s,\Delta_{\max}\} + 2\sqrt{2\sigma^2\frac{g_t(n_t^*)}{N_{s-1}^*}}
\: .
\end{align*}
We bound the divergence by $C$ and get
\begin{align*}
c_4(t)
&\leq \max_{k\neq *}\sum_{s\leq t, E_s^{*\%2}\cap\mathcal{E}_{n_{s-1}^*}^{exp}}
C \left( \frac{\max\{\varepsilon_1,\Delta_{\max}\}}{\Delta_{\min}} + \sqrt{\frac{8\sigma^2 g_t(n_t^*)}{(\Delta_s^k)^2 N_{s-1}^*}}\right)
\sqrt{\frac{8\sigma^2 g_t(n_t^*)}{(\Delta_s^k)^2 N_{s-1}^*}}\\
&\leq \sum_{s\leq t, E_s^{*\%2}\cap\mathcal{E}_{n_{s-1}^*}^{exp}}
C \left( \frac{\max\{\varepsilon_1,\Delta_{\max}\}}{\Delta_{\min}} + \sqrt{\frac{8\sigma^2 g_t(n_t^*)}{\Delta_{\min}^2 N_{s-1}^*}}\right)
\sqrt{\frac{8\sigma^2 g_t(n_t^*)}{\Delta_{\min}^2 N_{s-1}^*}}
\: .
\end{align*}

By design of the algorithm, $N_{s-1}^* \geq n_{s-1}^*/2$.
\begin{align*}
c_4(t)
&\leq 2C\frac{\max\{\varepsilon_1,\Delta_{\max}\}}{\Delta_{\min}^2}
\sum_{s\leq t, E_s^{*\%2}\cap\mathcal{E}_{n_{s-1}^*}^{exp}} \sqrt{2\sigma^2\frac{g_t(n_t^*)}{ n_{s-1}^*}}
+ \frac{8\sigma^2 C}{\Delta_{\min}} \sum_{s\leq t, E_s^*\cap\mathcal{E}_{n_{s-1}^*}^{exp}} \frac{g_t(n_t^*)}{ n_{s-1}^*}\\
&\leq 4C\frac{\max\{\varepsilon_1,\Delta_{\max}\}}{\Delta_{\min}^2} \sqrt{2\sigma^2(n_t^* - m_t^*) g_t(n_t^*)} + \frac{8\sigma^2 C}{\Delta_{\min}} \log(n_t^* - m_t^*) g_t(n_t^*)
\: .
\end{align*}

The definition of $C_5(t)$ is
\begin{align*}
C_5(t)
&= 2c_0^*(t) + \frac{1}{D}\left[c_4(t) + R_{n_t^*/2}^w + R_{n_t^*/2}^\lambda +  3L c_0^*(t) + Cm_t^*\right] \: .
\end{align*}
\end{proof}

\begin{lemma}
Let $a,b>0$. Let $x>0$ be such that $x \le a + \sqrt{x \log(bx)}$. Then for all $\eta\in(0,1)$
\begin{align*}
x \le \max\left\{\frac{a}{1- \eta}, \frac{1}{\eta^2} \overline{W}(\log \frac{b}{\eta^2})\right\} \: .
\end{align*}
\end{lemma}

\begin{proof}
The two terms of the max correspond to the two possibilities $\log(bx)<\eta^2 x$ and $\log(bx)\ge \eta^2 x$.
\end{proof}


\section{Perturbation}\label{app:perturbation.cost}

\begin{proof}[Proof of Theorem~\ref{thm:perturbation.cost}]

  Fix $\vmu$, $c>0$ and a small $\epsilon \le \Delta_{\min}$. Let $*$ denote the best arm for $\vmu$, and let $\q$ be an optimal solution for problem $D_\epsilon^{\mathcal M}(\vmu)$ from \eqref{eq:dual} which is supported on $\partial_\vmu \neg i^*(\vmu)$. Then in particular, with the sub-Gaussian property (Assumption~\ref{ass:sub_gaussian}),
  \[
    \frac{\ex_{\vlambda \sim \q} (\mu^* - \lambda^*)^2}{2 \sigma^2 \epsilon}
    ~\le~
    \frac{\ex_{\vlambda \sim \q}\sbr*{d(\mu^*,\lambda^*)}}{\epsilon}
    ~\le~
    D_\epsilon^{\mathcal M}(\vmu)\: .
  \]
Consider any $\vlambda \in \partial_\vmu\neg i^*(\vmu)$. By Assumption~\ref{ass:neg_is_lip}, we know that unless $\neg_{\mu^*}* = \emptyset$ there is another $\tilde \vlambda \in \neg *$ such that $\tilde\lambda^* = \mu^*$, and moreover for all $k\in[K]$, $\vert \tilde \lambda^k - \lambda^k \vert \le c_{\mathcal M} \vert \lambda^* - \mu^* \vert$. If $\neg_{\mu^*}* = \emptyset$ then $D^{\mathcal M}(\vmu) = 0$ and the inequality of the theorem is verified.

Let $L'>0$ be such that for all $k\in[K]$, $x\mapsto d(\mu^k, x)$ is $L'$-Lipschitz (Lemma~\ref{lem:bounds_on_d}). Then
\begin{align*}
d(\mu^k, \tilde \lambda^k)
&\le d(\mu^k, \lambda^k) + L' \vert \lambda^k - \tilde \lambda^k \vert
\\
&\le d(\mu^k, \lambda^k) + L' c_{\mathcal M} \vert \lambda^* - \mu^* \vert \: ,
     \intertext{Let $\tilde\q$ be any distribution obtained from $\q$ by replacing each $\vlambda$ in its support by such a $\tilde \vlambda$. Then
     }
\mathbb{E}_{\tilde \vlambda \sim \tilde \q} d(\mu^k, \tilde \lambda^k)
&\le \mathbb{E}_{\vlambda \sim \q} d(\mu^k, \lambda^k) + L' c_{\mathcal M} \mathbb{E}_{\vlambda \sim \q} \vert \lambda^* - \mu^* \vert
\\
&\le \mathbb{E}_{\vlambda \sim \q} d(\mu^k, \lambda^k) + L' c_{\mathcal M} \sqrt{\mathbb{E}_{\vlambda \sim \q} ( \lambda^* - \mu^* )^2}
\\
&\le \mathbb{E}_{\vlambda \sim \q} d(\mu^k, \lambda^k) + L' c_{\mathcal M} \sqrt{2 \sigma^2 \varepsilon D^{\mathcal M}_{\varepsilon}(\vmu)}
\: .
\end{align*}
This implies in particular that,
\begin{align*}
    D^{\mathcal M}(\vmu)
    &~\le~
      \max_{k \neq *}
    \frac{\ex_{\tilde\vlambda \sim \tilde\q}\sbr*{d(\mu^k,\tilde\lambda^k)}}{\Delta^k}
    \\
    &~\le~
      \max_{k \neq *}
    \frac{\mathbb{E}_{\vlambda \sim \q} d(\mu^k, \lambda^k) + L' c_{\mathcal M} \sqrt{2 \sigma^2 \varepsilon D^{\mathcal M}_{\varepsilon}(\vmu)}}
      {\Delta^k}
    \\
    &~\le~
    D_\epsilon^{\mathcal M}(\vmu)
    +
    \frac{L' c_{\mathcal M}\sqrt{2 \sigma^2 D_\varepsilon^{\mathcal M}(\vmu)}}
        {\Delta_{\min}}
    \sqrt{\varepsilon}
    \: .
\end{align*}
We may invert this using that $x \le y + c \sqrt{y}$ with $c \ge 0$ implies that $y \ge x - \frac{c}{2} \del*{\sqrt{c^2+4 x}-c} \ge x - c \sqrt{x}$, and hence we find
\[
  D_\epsilon^{\mathcal M}(\vmu)
  ~\ge~
  D^{\mathcal M}(\vmu)
  - \frac{L' c_{\mathcal M}\sqrt{2 \sigma^2 D^{\mathcal M}(\vmu)}}
  {\Delta_{\min}}
  \sqrt{\varepsilon}
  .
  \qedhere
\]
\end{proof}


\section{Example of Structures: Verifying Assumption~\ref{ass:neg_is_lip}}\label{app:verify_assumption}

In the following examples, we will describe possibly unbounded structure sets $\mathcal U \subseteq \Theta^K$. Then our algorithm will apply to $\mathcal M = \mathcal U \cap \{\Delta_{\min} > 0\} \cap [a,b]^K$ for an interval $[a,b] \subseteq \Theta$ such that $\mathcal M$ verifies Assumptions~\ref{ass:i*_unique} and~\ref{ass:M_bounded}.

\paragraph{Categorised, Lipschitz, Unconstrained or Unimodal.}

Unconstrained: $\mathcal U = \Theta^K$.

Unimodal: $\mathcal U = \{\vtheta \in \Theta^K : \exists k_0 \in [K], \theta^1 \le \theta^2 \le \ldots \le \theta^{k_0} \ge \ldots \ge \theta^K\}$.

Lipschitz: $\mathcal U = \{\vtheta \in \Theta^K : \forall k\in[K-1], |\theta^k - \theta^{k+1}| \le L\}$.

Categorised: the arms are known to each belong to one of $C$ categories, and one of the categories $c^*$ verifies that $\min_{k\in c^*}\theta^k$ is larger than $\max_{k\in c}\theta^k$ for all other categories.

We verify that Assumption~\ref{ass:neg_is_lip} holds with $c_{\mathcal M} = 1$, using the observation that for these structures, translating all arms by the same amount keeps the structure.
Set $\vmu \in \mathcal M$ and let $* = i^*(\vmu)$. Let $\vlambda \in \neg *$. Let $\vxi$ be defined by $\xi^k = \max\{a, \min\{\lambda^k + (\mu^* - \lambda^*) , b\}\}$. Then $\vxi \in \neg_{\mu^*}*$ and for all $k\in[K]$, $\vert \xi^k - \lambda^k \vert \le \vert \mu^* - \lambda^* \vert$. QED.

\paragraph{Linear.}

A set of arm vectors $a_1,\ldots, a_K \in \mathbb{R}^d$ is given. let $A \in \mathbb{R}^{K\times d}$ be the matrix with rows $a_1^\top, \ldots, a_K^\top$. The structure is $\mathcal U = \{\vlambda \in \Theta^K : \exists \veta \in \mathbb{R}^d, \vlambda = A \veta\}$.

We verify that Assumption~\ref{ass:neg_is_lip} holds.
Set $\vmu \in \mathcal M$ and let $* = i^*(\vmu)$. Let $\vlambda \in \neg *$. 
%
%
%
%
Then let $\veta \in \mathbb{R}^d$ be such that $\vlambda = A \veta$ and $\vzeta$ be a solution of the following problem, if there exists one:
\begin{align*}
\inf_{\vzeta \in \mathbb{R}^d} &\max_k \vert a_k^\top(\veta - \vzeta) \vert \\
\text{s.t. } & a_*^\top \vzeta = \mu^*, \\&(a_j - a_*)^\top \vzeta \ge 0
\: .
\end{align*}
Remark that it verifies
\begin{align*}
\left\{
  \begin{array}{ll}
    \displaystyle
\inf_{\vzeta \in \mathbb{R}^d} &\max_k \vert a_k^\top(\veta - \vzeta) \vert \\
\text{s.t. } & a_*^\top \vzeta = \mu^*, \\&(a_j - a_*)^\top \vzeta \ge 0
\end{array}
\right.
~\le~
\left\{
  \begin{array}{ll}
        \displaystyle
\inf_{\u \in \mathbb{R}^d} &\max_k \vert a_k^\top \u \vert \\
\text{s.t. } & a_*^\top \u = \mu^*-\lambda^*, \\&(a_j - a_*)^\top \u \ge 0
\end{array}
\right.
~\le~
\vert \mu^* - \lambda^* \vert
\left\{
  \begin{array}{ll}
        \displaystyle
\inf_{\u \in \mathbb{R}^d} &\max_k \vert a_k^\top \u \vert \\
\text{s.t. } & a_*^\top \u = 1, \\
& a_j^\top \u \ge 1 \text{ if } \mu^* > \lambda^*\\
& a_j^\top \u \le 1 \text{ if } \mu^* < \lambda^*
\end{array}
\right.
\end{align*}
and a solution $\u$ to the final optimisation problem gives a feasible $\vzeta = \veta + (\mu^* - \lambda^*) \u$ for the first problem. If $a_*$ and $a_j$ are not collinear, a solution exists. If all arms are collinear, then $\neg_{\mu^*}* = \emptyset$. Otherwise, there exists at least one suitable $j$.

For $a_j$ not collinear with $a_*$, let $c_{j,*}$ be the maximal value of the last optimisation problem for the two possibilities $\mu^*>\lambda^*$ and $\mu^*<\lambda^*$. Then $\Vert \vlambda - A \vzeta \Vert_\infty \le \vert \lambda^* - \mu^* \vert c_{j,*}$.

Assumption~\ref{ass:neg_is_lip} is verified for $c_{\mathcal M} = \max_{i,j \in [K]} c_{i,j}$ where $c_{i,j}$ is taken to be 0 if $a_i$ and $a_j$ are collinear.

\paragraph{Sparse.}

Among $K$ arms, $s$ arms ($s$ is known) have means greater than a known level $\gamma \in \Theta$. The $K-s$ other arms have mean $\gamma$.
Let $S_\vmu$ be the sparse support of $\vmu$, i.e. the set of $s$ arms with values greater than $\gamma$.

\begin{lemma}\label{lem:sparse_decomposition}
For all $\vmu \in \mathcal M$,
\begin{align*}
\partial_\vmu \neg i^*(\vmu)
\subseteq &(\cup_{j\in S_\vmu, j\neq *} \{\vxi \in \Theta^K \::\: \xi^j=\xi^*,\forall k \notin\{*,j\}, \xi^k = \mu^k\})
	\\
	&\cup
	(\cup_{j \notin S_\vmu}
		(\cup_{i\in S_\vmu,i\neq *}\{\vxi \in \Theta^K \::\: \xi^i=\gamma, \xi^j=\xi^*;\forall k \notin\{*,i,j\}, \xi^k = \mu^k\}
		\\
		& \qquad \qquad \cup \{\vxi \in \Theta^K \::\: \xi^*=\gamma;\forall k \neq *, \xi^k = \mu^k\}
		)
	)
	\: .
\end{align*}
\end{lemma}
\begin{proof}
Let $\vxi \in \neg *$ with best arm $j$ (or one of the best arms being $j$). If $j$ is in the sparse support of $\vmu$, then define $\vlambda$ equal to $\vmu$ except that $\lambda^* = \xi^*$ and $\lambda^j = \xi^j$. Then $\vlambda \in \neg *$ since it is $s$-sparse and $\lambda^j\ge \lambda^*$. Remark then that $\vxi \in Q_\vmu(\vlambda)$. If $\lambda^j<\mu^*$, let $\vlambda_2$ be equal to $\vlambda$ except that $\lambda_2^*=\lambda^j$, otherwise let $\vlambda_2$ be equal to $\vlambda$ except that $\lambda_2^j=\lambda_2^*=\mu^*$. Then $\vlambda_2$ belongs to the described set and $\vlambda \in Q_\vmu(\vlambda_2)$.

If $j$ is not in the sparse support of $\vmu$, let $i$ be in that support such that $\xi^i = \gamma$ (it exists since $\vxi$ is also $s$-sparse). Define $\vlambda$ with coordinates equal to those of $\vmu$ except for $*,i,j$, for which they are equal to the coordinates of $\vxi$. Then $\vlambda \in \neg *$ since it is $s$-sparse and $\lambda^j\ge \lambda^*$. Then $\vxi \in Q_\vmu(\vlambda)$.

If $i = *$, then let $\vlambda_2$ be the same as $\vlambda$ except that $\lambda_2^j = \mu^j$. $\vlambda \in Q_\vmu(\vlambda_2)$ and $\vlambda_2$ belongs to the prescribed set. Otherwise let $\vlambda_2$ be equal to $\vlambda$ except that if $\lambda^j<\mu^*$, $\lambda_2^* = \lambda^j$ and if $\lambda^j\ge \mu^*$, $\lambda_2^j = \lambda_2^* = \mu^*$. Then $\vlambda_2 \in \neg *$ and belong to one of the sets in the decomposition and $\vlambda \in Q_\vmu(\vlambda_2)$.
\end{proof}

\begin{lemma}
Assumption~\ref{ass:neg_is_lip} holds for the Sparse structure with $c_{\mathcal M}=1$.
\end{lemma}
We note that the assumption holds trivially for $s=1$, since in that case $\neg_{\mu^*}*$ is empty for all $\vmu \in \mathcal M$.
\begin{proof}
We show that the assumption holds for each of the sets in the decomposition of Lemma~\ref{lem:sparse_decomposition}.

\textbf{Case 1: $j\in S_\vmu$.} Let $\vlambda \in \neg * \cap\{\vxi \in \Theta^K \::\: \xi^j=\xi^*,\forall k \notin\{*,j\}, \xi^k = \mu^k\})$. Then $\vlambda + (\mu^*-\lambda^*)(\e_j + \e_*) \in \neg_{\mu^*}*$ and it verifies the conditions of the assumption with $c_{\mathcal M}=1$.

\textbf{Case 2: $j \notin S_\vmu$.} Let $i\in S_\vmu, i\neq *$ and $\vlambda \in \neg * \cap \{\vxi \in \Theta^K \::\: \xi^i=\gamma, \xi^j=\xi^*;\forall k \notin\{*,i,j\}, \xi^k = \mu^k\}$. Then $\vlambda + (\mu^*-\lambda^*)(\e_j + \e_*) \in \neg_{\mu^*}*$ and it verifies the conditions of the assumption with $c_{\mathcal M}=1$.

Let $\vlambda \in \{\vxi \in \Theta^K \::\: \xi^*=\gamma;\forall k \neq *, \xi^k = \mu^k\}$. If $\neg_{\mu^*}* \neq \emptyset$, then $s>1$. Take $l,m\neq *$ with $\lambda^m>\gamma$, and remark that $\vlambda +(\mu^* - \gamma)\e_*+(\mu^* - \lambda^l)\e_l - (\lambda^m - \gamma)\e_m \in \neg_{\mu^*}*$ and it verifies the conditions of the assumption with $c_{\mathcal M}=1$. 
\end{proof}



\section{Experiments Notes}\label{sec:impldet}
The experiment code is available in the repository \url{https://bitbucket.org/wmkoolen/tidnabbil/src/master/regret_games_paper/}.

\subsection{Parameters used for the Competition}
We use OSSB with forced exploration parameter $\epsilon = 0.02/\sqrt{\ln T}$ and concentration threshold $f(t) = (1+\gamma)\ln(1+t)$ with explore-exploit threshold parameter $\gamma = 0$. We run CATSE with $\delta=1/T$.

\subsection{Implementation Notes}
We describe how to implement the required alt-min oracle \eqref{eq:altmin} for the structures used in our experiments. Recall that $k$ is the index of the best arm in bandit instance $\vmu$, and $j$ is the index of another arm, which we are required to make better than $k$ in $\vlambda$.
\begin{itemize}
\item For the unconstrained case, we set $\lambda_k = \lambda_j$ equal to the weighted average $\frac{N_j \mu_j + N_k \mu_k}{N_j + N_k}$, and we set $\lambda_i = \mu_i$ for all $i \notin \set*{k,j}$.
\item For the $s$-sparse case, we reason separately about the three possible combinations of $k$ and $j$ being sparse or not. If both are not sparse, we equalise them as in the unconstrained case. For the other arms $i \notin \set*{k,j}$, we set $\lambda_i = \gamma$ for the arms minimising $N_i d(\mu_i, \gamma)$, and we set $\lambda_i = \mu_i$ for the remaining arms.
\item For the categorised case, we distinguish the case where $k$ and $j$ are in the same category, and the case where they are not. We have only implemented the case of two categories. Our approach is to find a separating level between the two categories by binary search.
\item For the Lipschitz, Simplex and Concave case we use the general-purpose convex quadratic optimiser OSQP through the JuMP interface.
\item For the unimodal case we use the PAVA algorithm for isotonic regression. We perform left-to-right increasing regressions and right-to-left decreasing regression, storing all the partial optima. This allows us to answer for any $k \neq j$ the cost of increasing up to arm $k$ and then decreasing from $k+1$ to the end.
\item For the linear case we first try the unconstrained projection, which has a linear algebra closed form. If this leaves $\lambda_k$ and $\lambda_j$ in the wrong order, we add the equality constraint and perform the projection in one dimension less.
\end{itemize}


\end{document}